\documentclass[10pt,twocolumn,letterpaper]{article}

\usepackage[pagenumbers]{cvpr} % To force page numbers, e.g. for an arXiv version

\newcommand{\na}{\textemdash}
\usepackage{mathtools}
\usepackage{booktabs,makecell,colortbl,xcolor,pifont}
\usepackage{wrapfig,lipsum}
\usepackage{wrapfig,lipsum}
\usepackage{tabularx}
\usepackage[accsupp]{axessibility}  % Improves PDF readability for those with disabilities.
\definecolor{pnpgray}{RGB}{120,120,120}
\definecolor{bestblue}{RGB}{0,92,184}
\definecolor{secondred}{RGB}{200,0,0}
\definecolor{secondred}{RGB}{200,0,0}
\definecolor{MyGreen}{RGB}{0,200,0}
\definecolor{MyBlue}{RGB}{0,0,200}

\definecolor{teaserColor1}{RGB}{248, 206, 204}
\definecolor{teaserColor2}{RGB}{218, 232, 252}

\usepackage{caption,booktabs}
\usepackage{float}
\usepackage{array}
\usepackage{graphicx}
\usepackage{footmisc}
\newcommand{\best}[1]{\textbf{#1}}            % best within group
\newcommand{\second}[1]{\underline{#1}}       % 2nd within group
\usepackage{pifont}
\newcommand{\cmark}{\ding{51}}
\newcommand{\xmark}{\ding{55}}
\usepackage{multirow}

\definecolor{cvprblue}{rgb}{0.21,0.49,0.74}
\usepackage[pagebackref,breaklinks,colorlinks,allcolors=cvprblue]{hyperref}

\def\paperID{*****} % *** Enter the Paper ID here
\def\confName{CVPR}
\def\confYear{2026}

\def\methodname{\texttt{Cov2Pose}}
\title{\methodname: Leveraging Spatial Covariance for Direct Manifold-aware 6-DoF Object Pose Estimation
 }

\author{Nassim {Ali Ousalah}$^{\star}$, Peyman Rostami$^{\star}$, Vincent Gaudillière$^{\dagger}$\\ Emmanuel Koumandakis$^{\rtimes}$, Anis Kacem$^{\star}$, Enjie Ghorbel$^{\star \ddagger}$ and Djamila Aouada$^{\star}$\\
CVI$^2$, Interdisciplinary Centre for Security, Reliability and Trust (SnT), University of Luxembourg$^{\star}$\\
Université de Lorraine, CNRS, Inria, LORIA, F-54000 Nancy, France$^{\dagger}$, Infinite Orbits$^{\rtimes}$\\
Cristal Laboratory, ENSI, University of Manouba$^{\ddagger}$\\
{\tt\small \{nassim.aliousalah,peyman.rostami,anis.kacem,djamila.aouada\}@uni.lu}\\
{\tt\small enjie.ghorbel@isamm.uma.tn, manos@infiniteorbits.io, vincent.gaudilliere@loria.fr}
\vspace{-3mm}
}

\begin{document}
\maketitle
\begin{abstract}
In this paper, we address the problem of 6-DoF object pose estimation from a single RGB image. Indirect methods that typically predict intermediate 2D keypoints, followed by a Perspective-$n$-Point solver, have shown great performance. Direct approaches, which regress the pose in an end-to-end manner, are usually computationally more efficient but less accurate. However, direct pose regression heads rely on globally pooled features, ignoring spatial second-order statistics despite their informativeness in pose prediction. They also predict, in most cases, discontinuous pose representations that lack robustness.  Herein, we therefore propose a covariance-pooled representation that encodes convolutional feature distributions as a symmetric positive definite (SPD) matrix. Moreover, we propose a novel pose encoding in the form of an SPD matrix via its Cholesky decomposition. Pose is then regressed in an end-to-end manner with a manifold-aware network head, taking into account the Riemannian geometry of SPD matrices. Experiments and ablations consistently demonstrate the relevance of second-order pooling and continuous representations for direct pose regression, including under partial occlusion.
\vspace{-3mm}
\end{abstract}
    
\section{Introduction}
\label{sec:intro}
Estimating the six-degree-of-freedom (6-DoF) pose of an object from an RGB image is a core challenge in computer vision~\cite{Li_2025_CVPR, Lee_2025_CVPR, Huang_2025_CVPR, Tan_2025_CVPR, lian2025vapo, ousalah2025uncertainty, Liu_2025_CVPR, ousalah2025fpg} due to its wide range of applications in robotic manipulation~\cite{11092949, 10766628}, augmented and virtual reality~\cite{foundationposewen2024}, autonomous navigation~\cite{cassinis2019review, PAULY2023339}, and industrial inspection~\cite{10657978, Maack_2025_WACV, GOVI2024104130}. Given an image, the goal is to predict the pose $P \in \mathrm{SE}(3)$ of the object of interest relative to the camera that captured the image, where $\mathrm{SE}(3)$ denotes the Special Euclidean Group of $\mathbb R^3$.

\begin{figure}[t]
    \centering
    \includegraphics[width=0.45\textwidth]{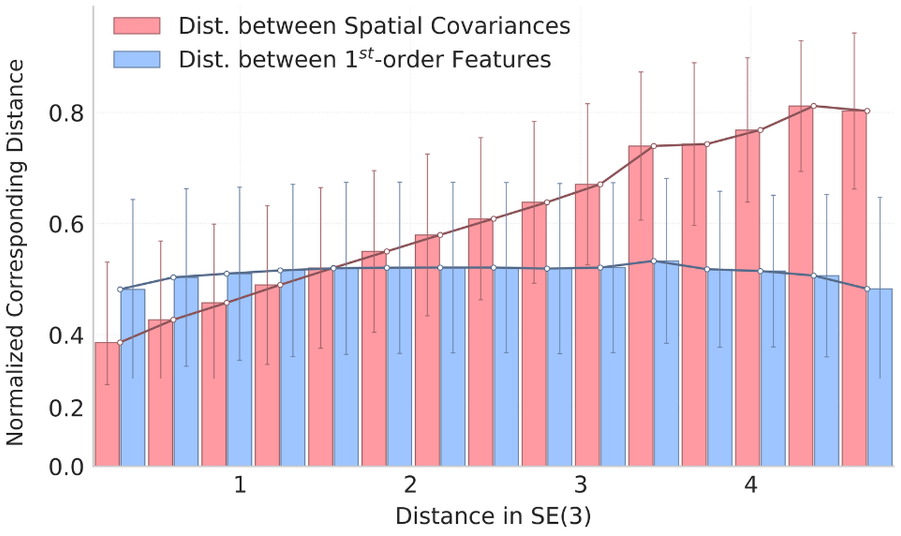}
    \caption{\textbf{Pose–representation alignment}. Pairs of images\protect\footnotemark{} are sorted according to the geodesic distances in $\mathrm{SE}(3)$ between their corresponding ground-truth poses (x-axis). The x-axis is then partitioned into bins. For each bin, we plot: (i) the average Log–Euclidean distance between spatial covariances of image pairs within this bin (red) and (ii) the average Cosine distance between flattened features of image pairs within the bin (blue). Covariance distances increase with pose distance, whereas flattened feature distances remain nearly constant, indicating that spatial covariance is more correlated with the pose.}
    \label{fig:motivation}
    % \vspace{-3mm}
\end{figure}
\footnotetext{Training images from the \texttt{002\_master\_chef\_can} class of the YCB-V~\cite{xiang2018posecnn} dataset. We use an EfficientNet-B6~\cite{tan2019efficientnet} pretrained on ImageNet~\cite{deng2009imagenet} with no prior training on pose estimation.}

Thanks to the tremendous developments made in deep learning, numerous approaches have leveraged Deep Neural Networks (DNNs)~\cite{xu20246d, jin20246dope, foundationposewen2024, lin2024sam} either directly or indirectly to estimate the 6-DoF object pose. Indirect methods mainly predict  intermediate representations such as keypoints~\cite{peng2019pvnet, xu20246d, wu2024learning, lian2025vapo} or dense object coordinates~\cite{zakharov2019dpod, wang2019normalized} that are used to solve a Perspective-$n$-Point (P$n$P) problem  or adopt a render-and-compare approach that optimizes photometric, silhouette, or differentiable rendering losses to align a rendered model view with the input image~\cite{li2018deepim, manhardt2018deep, labbe2020cosypose, li2024mrc}. Direct methods, on the other hand, regress rotation and translation in an end-to-end manner using a neural architecture~\cite{Li2019CDPNCD, xiang2018posecnn, wang2021gdr, do2018deep, wang2019densefusion, chen2022epro, 10937121}. While indirect methods have been demonstrated to be more accurate, they suffer from a high calculation cost induced by the iterative computation they incur~\cite{li2018deepim}. In contrast, direct methods are usually more efficient despite being less precise~\cite{wang2021gdr, chen2022epro, chen2020end}, thereby suggesting their practical relevance in real-time applications.

In this paper, we argue that we can improve the accuracy of direct methods by incorporating the spatial second-order statistics of the input image within the DNN-based pose estimator. In fact, deep pose regressors typically rely solely on first-order statistics of feature activations. Nonetheless, second-order moments such as covariance matrices can capture pairwise co-activations that average or max pooling operations discard, thereby providing information about the spatial variability within an image. We hypothesize that such information is highly relevant for inferring the pose. This hypothesis is confirmed by the experiment reported in Figure~\ref{fig:motivation}. This figure shows that pairwise distances between spatial covariances of DNN features increase with ground-truth $\mathrm{SE(3)} $ pose distance, whereas distances between flattened features remain nearly constant. This suggests that spatial covariance is strongly correlated with pose and provides, as a result, a more suitable representation for direct pose regression.

Concretely, we propose to make use of second-order pooling, which has shown its utility in several deep learning-based computer vision tasks such as classification, facial expression recognition, object detection, and action recognition~\cite{7410696, 6479703, gao2019global, NEURIPS2021_70efdf2e, 9001240, wang2020deep, 10.1007/978-3-642-33786-4_32, acharya2018covariance}. This pooling strategy returns feature covariance matrices that are known to be symmetric positive-definite (SPD), ensuring the exploitation of statistical information beyond first-order within a DNN, while preserving spatial awareness. To the best of our knowledge, we are the first to explore spatial covariance representations for end-to-end pose regression.

Our framework, termed \methodname, is an end-to-end trainable pipeline that respects the geometry of the SPD manifold where covariance matrices reside. Moreover, we introduce within \methodname~a one-to-one continuous mapping from the SPD space to $\mathrm{SE}(3)$ via a differentiable Cholesky-based encoding without breaking the chain rule, allowing the direct prediction of a continuous 6-DoF pose representation. In particular, the pose is predicted under the shape of a continuous 6D rotation parametrization coupled with a 3D translation. This chosen rotation representation has been shown to be more suitable for rotation learning~\cite{zhou2019continuity}. Experiments performed on several object pose benchmarks show the superiority of \methodname~compared to previous direct methods, demonstrating the relevance of the proposed principled framework.

\noindent In summary, our main contributions can be summarized as follows,
\begin{itemize}
  \item The first covariance-based deep learning framework for RGB-based end-to-end 6-DoF object pose regression, which leverages the spatial covariance of backbone features to encode higher-order statistics.
  
  \item A manifold-aware training pipeline that applies geometry-preserving dimensionality reduction through Bilinear Mapping layers (BiMap)~\cite{huang2017riemannian} to a compact SPD representation.
  
   \item A fully differentiable, CAD-free, one-to-one, and continuous pose regressor that maps the latent SPD matrix to a continuous 6D rotation representation and translation via a differentiable Cholesky decomposition.

  \item Extensive experiments on three object pose benchmarks, namely, LineMOD~\cite{10.1007/978-3-642-37331-2_42}, Occ-LineMOD~\cite{10.1007/978-3-319-10605-2_35} and YCB-Video~\cite{xiang2018posecnn}, showing that   \methodname~achieves state-of-the-art results as compared to direct regression methods.

\end{itemize}

The remainder of the paper is organized as follows: Section~\ref{sec:relatedwork} presents the related work on 6-DoF object pose estimation and the background. Section~\ref{sec:problemstatement} formulates the problem of direct pose regression. Section~\ref{sec:methodology} describes the proposed framework entitled \methodname. Section~\ref{sec:experiments} reports the conducted experiments and discusses the obtained results. Finally, Section~\ref{sec:conclusion} concludes this work, highlighting potential future directions.

\section{Related Work and Background}
\label{sec:relatedwork}

In this section, we first review recent literature on RGB-based 6-DoF object pose estimation, then discuss the use of covariance representations in vision tasks. We also provide preliminaries on the SPD manifold, where such representations naturally reside, and review deep learning methods operating on these manifolds.

\paragraph{RGB-based 6-DoF Object Pose Estimation.} The field has progressed along three complementary paradigms. \textbf{\textit{PnP‑based methods}} first establish 2D–3D correspondences and then solve a P$n$P problem~\cite{RANSACPnP}. Early keypoint-based works~\cite{7989233,Tekin_2018_CVPR,Rad_2017_ICCV} introduced a learnable regressor for pre-defined sparse object landmarks coordinates or heatmaps. PV‑Net~\cite{peng2019pvnet} proposed to leverage all pixels through a voting system for more robust regression. More recent approaches proposed to learn the keypoints~\cite{wu2024learning}, or to refine keypoint prediction with a diffusion‑driven framework~\cite{xu20246d}. A visibility‑aware mechanism has also been proposed to handle occlusions in \cite{lian2025vapo}. Dense‑correspondence variants aim to improve pose accuracy by predicting per‑pixel matches across the whole object surface. Early examples are DPoD~\cite{zakharov2019dpod} and the normalized object coordinate space method~\cite{wang2019normalized}. More recently, ZebraPose~\cite{su2022zebrapose} introduced a coarse‑to‑fine surface encoding and CheckerPose~\cite{lian2023checkerpose} adopted progressive dense keypoint localisation. However, these methods generally suffer from a costly iterative outlier rejection process. \textbf{\textit{Render‑and‑compare methods}} iteratively render the object under a hypothesised pose and refine it against the observed image. The earliest contribution, DeepIM~\cite{li2018deepim}, introduced an iterative refinement scheme that directly optimises the pose by minimising photometric error. Manhardt \textit{et al.} extended this idea to a model‑based RGB refinement pipeline~\cite{manhardt2018deep}. RePose~\cite{Iwase_2021_ICCV} later demonstrated fast pose refinement via deep texture rendering. Subsequently, CosyPose added multi‑view consistency to improve robustness~\cite{labbe2020cosypose}, and MRC‑Net further enhanced the matching‑refinement cycle with multiscale residual correlation~\cite{li2024mrc}. However, these methods require a 3D object model to render and rely on an expensive iterative optimization loop. \textbf{\textit{Direct pose‑regression methods}} aim to predict the 6-DoF object pose in a single forward pass. Early end‑to‑end regressors such as PoseCNN~\cite{xiang2018posecnn} and Deep‑6DPose~\cite{do2018deep} demonstrated that a CNN can output pose parameters directly from RGB cues. Subsequent work introduced more structured regressors: CDPN employed a coordinates‑based disentangled network~\cite{Li2019CDPNCD}, DenseFusion fused RGB and depth features before regression~\cite{wang2019densefusion}, and GDR‑Net presented a geometry‑guided direct regression architecture~\cite{wang2021gdr}, extended in \cite{10937121}. Probabilistic P$n$P layers were incorporated in EPro‑PnP to model uncertainty and enable end-to-end training through intermediate 2D-3D correspondence predictions~\cite{chen2022epro}. Most prior methods overlook higher-order statistics such as covariance matrices, which, as shown in Figure~\ref{fig:motivation}, carry valuable information for pose regression. These neglected statistics are correlated with object poses and thus offer potential for improving direct pose estimation. Furthermore, existing approaches often rely on non-continuous rotation representations, which, as demonstrated in prior studies~\cite{zhou2019continuity}, hinder stable and effective pose learning.

\vspace{-0.5cm}
\paragraph{Covariances and Deep Learning on the SPD Manifold.} Covariance descriptors have long been employed to encode visual information across a wide range of computer vision tasks~\cite{11744047_45,AYEDI20121902,otberdout2018deep}. By aggregating second‑order statistics of local features, covariance matrices capture complementary texture, shape, and color cues while remaining compact and robust to noise. Because these descriptors are symmetric positive‑definite, they reside on the Riemannian manifold of SPD matrices. Formally, this manifold is defined as,
\begin{equation}
    \begin{aligned}
    \mathcal{S}^n_{++}
    = \big\{\,\mathbf M\in\mathbb{R}&^{n\times n}:\; \mathbf M=\mathbf M^{\top},\\
    & \mathbf x^{\top}\mathbf M \mathbf x>0\ \ \forall \mathbf x\in\mathbb{R}^n\setminus\{0\}\,\big\},
    \end{aligned}
\end{equation}
\noindent where ${}^\top$ denotes matrix transpose. This space lacks the linear structure of a vector space. Consequently, standard operations such as linear combinations are not directly applicable, as they do not necessarily preserve SPD nature. As such, algorithms should be tailored to respect the intrinsic non‑linear geometry of the manifold. Since standard convolutional or fully‑connected layers presume a Euclidean feature space, researchers have introduced SPD‑aware building blocks to embed the geometry into end‑to‑end trainable pipelines~\cite{huang2017riemannian,NEURIPS2019_6e69ebbf,Gao_2020_CVPR,Wang_2022_ACCV}. These SPD deep networks have demonstrated state‑of‑the‑art performance in tasks ranging from face~\cite{Dong_Jia_Zhang_Pei_Wu_2017} and action recognition~\cite{Nguyen_2019_CVPR} to medical image analysis~\cite{mayr2025covariancedescriptorsmeetgeneral}. Nevertheless, most of them have targeted classification tasks, while regression problems, such as pose estimation, have been overlooked. In this paper, we are the first to exploit SPD-aware deep learning for pose estimation. We briefly recall below the core contributions that have led to the development of SPD Deep Learning frameworks:

\begin{itemize}
    \item \textbf{\textit{BiMap (Bilinear Mapping).}} 
Similar to projection layers in Euclidean nets, a BiMap layer compresses the representation to generate more compact and discriminative outputs. With a column-orthonormal weight matrix 

$\mathbf W\!\in\!\mathcal{V}_n(\mathbb{R}^m)=\{\mathbf W\in\mathbb{R}^{n\times m}\mid \mathbf W^\top \mathbf W=\mathbf I_m\}$ ($\mathbf I_m$ being the identity matrix of size $m$, and $\mathcal{V}_n(\mathbb{R}^m)$ denotes the Stiefel manifold), it transports the input covariance by congruence,
\begin{equation}
\label{eq:BiMap}
    \mathbf Y=\mathbf W \mathbf X\,\mathbf W^\top,\qquad \mathbf X\in\mathcal S^n_{++},\ \mathbf Y\in\mathcal S^m_{++} \ ,
\end{equation}
This contracts $n\!\to\!m$ ($m<n$) while preserving the SPD structure. During training, $\mathbf W$ is kept on the Stiefel manifold~\cite{huang2017riemannian} via a QR retraction at each update.

\item \textbf{\textit{ReEig (Eigenvalue Rectification).}} 
As for ReLU in ConvNets, ReEig introduces nonlinearity while preserving the SPD structure by lifting small eigenvalues to a spectral floor. Let the $k$-th layer input be $\mathbf X\in\mathcal S_{++}^n$ with eigendecomposition $\mathbf X=\mathbf U\mathbf\Sigma\mathbf U^\top$. This operation is defined such that
\begin{equation}
\label{eq:ReEig}
\operatorname{ReEig}_\varepsilon(\mathbf X)=\mathbf U\,\max(\mathbf \Sigma,\varepsilon \mathbf I)\,\mathbf U^\top,\ \quad \varepsilon>0 \ ,
\end{equation}
where $\max$ acts element-wise on the diagonal, $\varepsilon$ sets the floor, and $\mathbf I$ is the identity matrix. This keeps the output in $\mathcal S_{++}^n$ and prevents the collapse of small modes.
\end{itemize}

\begin{figure*}[t]
    \centering
    \includegraphics[width=.99\textwidth]{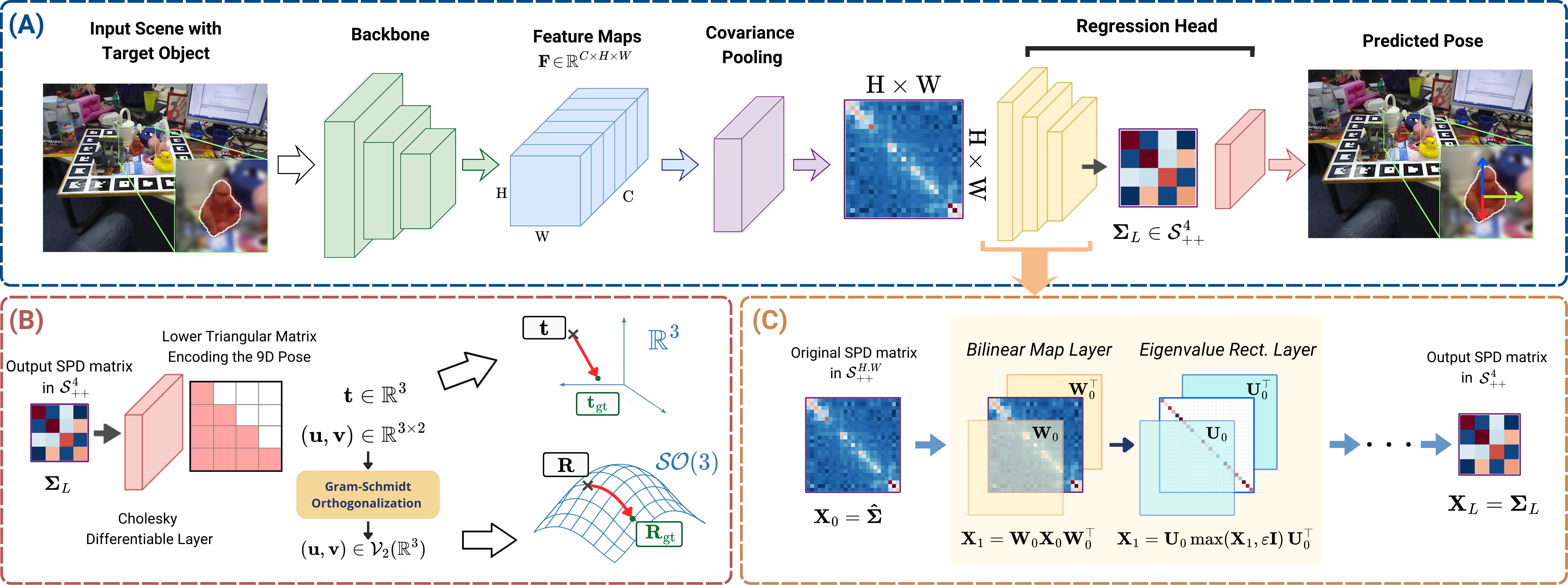}

    \caption{\textbf{Overview of the proposed pose estimation framework} (best viewed in color). \textbf{(A) General overview.} A CNN backbone extracts feature maps $\mathbf F\!\in\!\mathbb{R}^{C\times H\times W}$. Spatial covariance pooling over the $H{\times}W$ grid gives $\boldsymbol{{\hat \Sigma}}\!\in\!\mathcal{S}^{H{\times}W}_{++}$ which is fed  to the pose head. \textbf{(B) Learning the pose.} After $L$ geometry-aware layers, the compact SPD matrix $\mathbf \mathbf{\mathbf \Sigma_L}\!\in\!\mathcal{S}^{4}_{++}$ is decoded by a differentiable Cholesky layer into $(\mathbf u,\mathbf v;\mathbf t)$; $(\mathbf u,\mathbf v)$ are mapped to $\mathcal{SO}(3)$ and the loss is computed on $\mathcal{SO}(3)\times\mathbb{R}^3$. \textbf{(C) SPD learning layers.} Each layer applies a BiMap followed by eigenvalue rectification $\operatorname{ReEig}_{\varepsilon}$~\cite{huang2017riemannian} preserving strict positive definiteness.}
    \label{fig:method}
    \vspace{-2mm}
\end{figure*}

\section{Problem Statement}
\label{sec:problemstatement}

Estimating the 6-DoF pose of a given object from a sample \( \mathbf{I} \in \mathcal{I} \) belonging to the space of RGB images $\mathcal{I}$  involves finding a mapping \( \mathbf{\Phi} \) such that,
\begin{equation}
\label{eq:pose}
    \begin{array}{ccccll}
    \renewcommand{\arraystretch}{0.2}
        \mathbf{\Phi} ~~  : & \mathcal{I} & \longrightarrow &  \mathcal{P} \\
         & \mathbf I & \longmapsto & \mathbf{\Phi}(\mathbf I) & = & \mathbf P \ ,
    \end{array}
\end{equation}

\noindent where $\mathbf P \in \mathcal{P}$ is the pose of the object with respect to the camera coordinate system and $\mathcal{P}$ is the pose representation space. Given a set of training samples $\mathcal D=\{\mathbf I_i, \mathbf P_i\}_{i=1}^D$, where $\mathbf I_i  \in \mathcal{I}$ represents an image sample and  $\mathbf P_i \in \mathcal{P}$ its associated pose, direct methods typically train a neural network $\mathbf{\Phi}_\theta$ parameterized by weights $\theta$ to minimize a pose loss denoted as $ \mathcal L_{\text{pose}}$, 

\begin{equation}
\label{eq:regression}
\theta^\star=\underset{\theta}{\operatorname {arg\,min}}\ \sum_{i=1}^{D} \mathcal L_{\text{pose}}\!\left(\boldsymbol{\Phi}_\theta(\mathbf I_i),\mathbf P_i\right) \ .
\end{equation}

\noindent In turn, the learned neural network $\mathbf{\Phi}_\theta$  can be seen as the composition of two functions $\mathbf{\Phi}_\theta=\mathbf{\Psi}_{\theta_2} \circ \mathbf{\Gamma}_{\theta_1}$ where $\mathbf{\Gamma}_{\theta_1}:\mathcal{I}\longmapsto \mathcal{F} $ is the feature extractor that maps an input image into the feature space $\mathcal{F}$  and $\mathbf{\Psi}_{\theta_2}:\mathcal{F}\longmapsto \mathcal{P}$ that maps extracted features to the object pose. End-to-end methods mainly use: (\textit{i}) standard features that are extracted by DNNs using $\mathbf{\Gamma}_{\theta_1}$, ignoring second-order statistics that can be highly correlated with the pose, as suggested by the experiment presented in Figure~\ref{fig:motivation}; and (\textit{ii}) a non-continuous representation space $\mathcal{P}$, as regressing continuous representations using $\mathbf{\Psi}_{\theta_2}$ necessitates a manifold-aware learning strategy that is not straightforward to implement. In fact, most direct approaches regress the rotation either using quaternions or Euler angles that are non-continuous~\cite{zhou2019continuity}, inducing the non-continuity of the pose. Nevertheless, as demonstrated in \cite{geist2024learning}, the use of continuous representations via Gram-Schmidt orthonormalization or Singular Value Decomposition (SVD) is more suitable for learning rotations. This paper aims at addressing the two aforementioned issues by imposing the extraction of second-order statistics using $\mathbf{\Gamma}_{\theta_1}$ and constraining that $\mathbf{\Psi}_{\theta_2}$ maps to continuous pose representations in an end-to-end way.

\section{Proposed Approach}
\label{sec:methodology}

To address the problem outlined in Section~\ref{sec:problemstatement}, we introduce \methodname, the first covariance-based framework for end-to-end 6-DoF object pose regression that directly outputs a continuous pose representation, namely, 6D representation for rotation coupled with a translation vector. In particular, the initial feature maps extracted from the input image using a CNN backbone are first encoded in a covariance matrix using a second-order pooling.  Since covariance matrices are SPD, trainable BiMap layers are employed to progressively reduce their dimensionality while taking into account the geometry of the SPD manifold. As such, the succession of these two steps forms a feature extraction  $\mathbf{\Gamma}:\mathcal{I}\longmapsto \mathcal{S}_{++}^n $ based on second-order statistics. Finally, a differentiable Cholesky layer is added to decompose the reduced covariance matrix into two triangular matrices. The obtained matrices encode, in their non-zero elements, the 6D representation of the rotation, as well as the translation vector, allowing a continuous and injective mapping $\mathbf{\Psi}:\mathcal{S}_{++}^n \longmapsto \mathcal{P}$. The full pipeline of our solution is illustrated in Figure~\ref{fig:method}. In the following, we start by describing  the covariance-based feature extractor  $\mathbf{\Gamma}$ (Section~\ref{subsec:cov_pose_estim}). We then depict the proposed differentiable mapping $\mathbf{\Psi}$ from the resulting SPD matrix to the pose using Cholesky decomposition (Section~\ref{subsec:encoding}). Finally, the training strategy is presented in (Section~\ref{subsec:training}).

\subsection{Covariance-based Feature Extraction}
\label{subsec:cov_pose_estim}

\begin{figure}[t]
    \centering
    \includegraphics[width=0.49\textwidth]{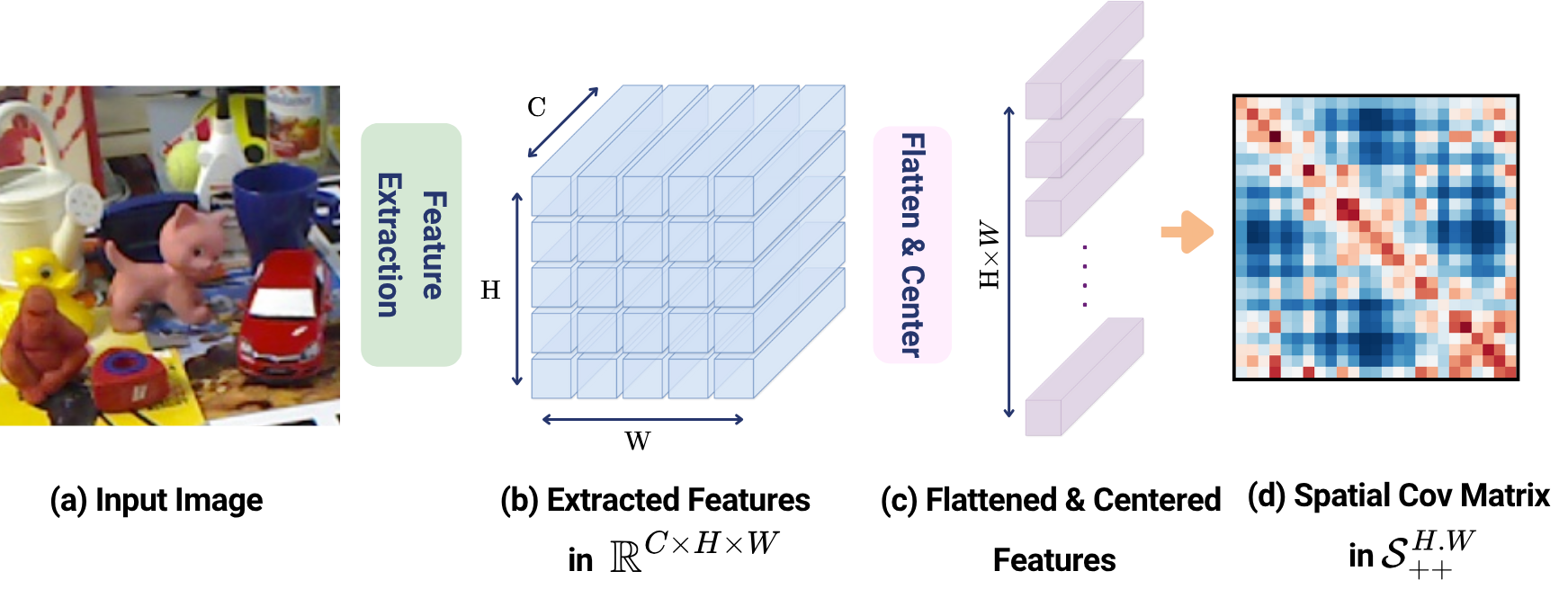}
    \caption{\textbf{Spatial Covariance Pooling.} Extracted features are flattened and centered and $2^{\text{nd}}$ order statistics are constructed pixel-wise. Every entry of the covariance matrix describes how two spatial locations of the given input co-variate.}
    \label{fig:covpooling}
    \vspace{-4mm}
\end{figure}

We first extract from an input image $\mathbf{I}$ of the scene, containing the target object, a set of feature maps \mbox{$\mathbf{\Gamma}_1(\mathbf{I})=\mathbf{F}\in\mathbb{R}^{C\times H\times W}$} using a CNN-based feature extractor $\mathbf{\Gamma}_1:\mathcal{I} \longmapsto \mathbb{R}^{C
{\times}
H
{\times}W}$, where $C$, $H$, and $W$ are the number of channels, height, and width, respectively. To include the second-order statistics in our framework, we propose incorporating a second-order pooling layer that computes a spatial covariance matrix $\mathbf{\Gamma}_2:\mathbb{R}^{C \times H \times W} \longmapsto \mathcal{S}_{++}^N $ of the extracted features $\mathbf{F}$, with $N=H \times W$. In particular, the covariance matrix ${\boldsymbol{\hat \Sigma}}$ is estimated  as follows,

\begin{equation}
\label{eq:cov}
\begin{aligned}
\boldsymbol{\hat \Sigma}
  &= \mathbf{\Gamma}_2(\mathbf F) = \operatorname{CovPool}(\mathbf X) \\
  &= \frac{1}{C-1}\sum_{i=1}^{C}
     (\mathbf X_i-\boldsymbol{\mu}_\mathbf{X})^{\!\top}(\mathbf X_i-\boldsymbol{\mu}_\mathbf{X}) \ ,
\end{aligned}
\end{equation}

\noindent where $\mathbf X=\text{vec}( \mathbf F) \in \mathbb R^ {C \times N} $ corresponds to the flattened feature maps over the spatial dimensions and \mbox{$\mathbf{\mu}_{\mathbf X}=  \frac{1}{C} \sum_{i=1}^{C} \mathbf X_i$} such that $\mathbf X_i \in R^ {N} $ is the flattened feature map resulting from the $i^{th}$ channel.  Figure~\ref{fig:covpooling} illustrates the spatial covariance pooling. This covariance-based representation offers a key advantage in the context of 6-DoF pose estimation. It models how spatial locations ``co-vary'' across the feature maps, a statistic that changes systematically according to the image viewpoint. As discussed in Figure~\ref{fig:motivation}, a clear correlation between the spatial covariance and the pose exists as compared to standard feature maps, making them highly informative for direct pose regression.

To regress the pose, it is necessary to reduce beforehand the dimensionality of the resulting high-dimensional covariance matrix. Nonetheless, covariance matrices in~\eqref{eq:cov} lie on the Riemannian manifold of SPD matrices, $\mathcal{S}_{++}^N$. Hence, applying standard deep learning reduction layers is not adequate, as it does not account for their non-linear structure, resulting in non-SPD outputs. In fact, standard DNNs assume that the nature of the feature space is Euclidean. Consequently, considering deep dimensionality reduction layers specifically tailored to the SPD manifold is essential. For that reason, we propose to employ $L$ BiMap layers~\cite{huang2017riemannian}, which allow for reducing the dimension of the covariance matrix while maintaining its geometric structure, as presented in~\eqref{eq:BiMap}.
Moreover, a ReEig~\cite{huang2017riemannian} operation discussed in~\eqref{eq:ReEig} is also applied to avoid singularities, resulting in symmetric positive semi-definite matrices. Mathematically, the final covariance-based features $\mathbf{\Sigma}_L=\mathbf{\Gamma}_3(\mathbf{\Sigma})$ with $\mathbf{\Gamma}
_3: \mathcal{S}_{++}^N\longmapsto \mathcal{S}_{++}^n $ and $n<<N$ are obtained as follows,

\begin{equation}
\label{eq:spd_head_stack}
\begin{aligned}
&\mathbf{\Sigma}_{l+1} &=& \quad \operatorname{ReEig}_{\varepsilon}\!\big(\mathbf W_l^\top \mathbf X_l \mathbf W_l\big) \ , \\
&\mathbf{\Sigma}_0 &=& \quad \hat{\boldsymbol\Sigma} \ ,\\
\end{aligned}
\end{equation}
for $l \in \{ 0,\ldots,L-1 \}$. In summary, the covariance-based feature extractor can be written as  $\mathbf{\Gamma}=\mathbf{\Gamma_3} \circ \mathbf{\Gamma_2} \circ  \mathbf{\Gamma_1}$. In our experiments, we set $\varepsilon$ of the ReEig layer to $10^{-4}$.

\subsection{From SPD to Pose Parametrization}
\label{subsec:encoding}
Given a reduced covariance based descriptor \mbox{$\mathbf{\Sigma}_L=\mathbf{\Gamma}(\mathbf{I}) \in \mathcal{S}_{++}^n$} extracted from an image $\mathbf{I}$, the goal is to seek a mapping %$\mathbf{\Psi}$ :
\begin{equation}
    \begin{array}{cccccc}
        \mathbf{\Psi} & : & \mathcal{S}^n_{++} & \longrightarrow & \mathcal P \\
         & & \mathbf{\Sigma}_L & \longmapsto & \mathbf{\Psi} (\mathbf{\Sigma}_L) = \mathbf{P} \ , 
    \end{array}
    \label{eq:psi}
\end{equation}

\noindent with $\mathbf{P}$ being the pose associated with $\mathbf{I}$.

\vspace{0.1cm}
\noindent \textbf{Continuity of $\mathcal{P}$. } As discussed in Section~\ref{sec:problemstatement}, we favor picking $\mathcal{P}$ as a continuous space, which will define the nature of the used pose representation. The pose space $\mathcal{P} = \mathcal{R} \times \mathcal{T}$ can be seen as the direct product of the rotation space denoted as $\mathcal{R}$, and the position space $\mathcal{T}$. While the use of a simple vector in $\mathbb{R}^3$ for translation guarantees the continuity of $\mathcal{T}$, the continuity of rotation representations is not always ensured, as for Euler angles and quaternions~\cite{zhou2019continuity}. To ensure continuity, two common alternatives from the literature may be considered, namely the 6D representation with Gram-Schmidt orthonormalization and the 9D Special Orthogonal representation. In the following, we choose to employ the 6D representation with Gram-Schmidt orthonormalization given its relative simplicity. In fact, when learning the 6D representation, only two orthonormal vectors are learned. 

As such, the rotation representation space $\mathcal{R}$ is defined as the Stiefel manifold~\cite{Chen_2022_CVPR, absil2008optimization} denoted as $\mathcal{V}_2(\mathbb{R}^3)$ and therefore the pose space as $\mathcal{P} = \mathcal{V}_2(\mathbb{R}^3){\times} \mathbb{R}^3$.

\vspace{0.1cm}
\noindent \textbf{Properties of $\mathbf{\Psi}$. } The mapping $\boldsymbol{\Psi}$ defined in~\eqref{eq:psi} should be: 
\begin{enumerate}
    \item~\textbf{Injective}, providing a unique pose for a given SPD matrix input;
    \item~\textbf{Continuous}, ensuring that two close SPD matrices result in close poses;
   \item~\textbf{Differentiable}, enabling its incorporation within an end-to-end deep learning framework. 
\end{enumerate}%(1) (2)~\textbf{continuous}: for ensuring that two close covariance matrices result in close poses; (3)~\textbf{differentiable}: for enabling its incorporation within and end-to-end deep learning framework.

\noindent The Cholesky decomposition is known to be unique, continuous~\cite{Schatzman_Cholesky} and differentiable~\cite{murray2016differentiation, Seeger2017Differentiating}, making it an ideal candidate for defining $\mathbf{\Psi}$. Specifically, it decomposes any SPD matrix into the product of a lower triangular matrix and its conjugate transpose. As a result, the obtained triangular matrix can be used for encoding the selected 6D rotation representation (6-dimensional) as well as the translation vector (3-dimensional). For that purpose, a matrix with more than 9 non-zero elements is needed, necessitating fixing $n$ to 4.
This can be defined as $\boldsymbol{\Psi}(\boldsymbol{\Sigma}_L) = \mathbf{L}$, such that $\boldsymbol{\Sigma}_L = \mathbf{L}\mathbf{L}^\top \in \mathcal{S}^{4}_{++}$ and,

\begin{equation}
\label{eq:structured-cholesky}
% \resizebox{0.4\textwidth}{!}{$
\begin{aligned}
% \boldsymbol{\Psi}(\boldsymbol{\Sigma}_L) &= \mathbf{L}, 
% \quad \text{such that} \quad 
% \boldsymbol{\Sigma}_L = \mathbf{L}\mathbf{L}^\top \in \mathcal{S}^{4}_{++}, \\%[6pt]
\mathbf{L} &=
\begin{bmatrix}
e^{t_x} & 0 & 0 & 0\\
u_1 & e^{t_y} & 0 & 0\\
u_2 & v_1 & e^{t_z} & 0\\
u_3 & v_2 & v_3 & e^{-(t_x+t_y+t_z)}
\end{bmatrix} \ ,
\end{aligned}
% $}
\end{equation}

\noindent where $\mathbf u=(u_1,u_2,u_3)^\top$ and $\mathbf v=(v_1,v_2,v_3)^\top$, with $(\mathbf u,\mathbf v)~\in~\mathbb{R}^{3\times2}$, are the two 3D vectors that parameterize the 6D rotation representation and $\mathbf t=(t_x,t_y,t_z)$ is the translation vector. Following~\cite{zhou2019continuity,Chen_2022_CVPR}, we first map $(\mathbf{u}, \mathbf{v})$ onto the Stiefel manifold $\mathcal{V}_2(\mathbb{R}^3)$ using a differentiable Gram-Schmidt process, then construct the rotation matrix $\mathbf{R} \in \mathcal{SO}(3)$ by completing the basis with the cross product of the two orthonormalized vectors. Here, $\mathcal {SO}(3)$ denotes the Special Orthogonal group of $\mathbb {R}^3$. This map is continuous except on the set where $\mathbf u$ and $\mathbf v$ are collinear. 

Note that the exponential on the diagonal of $\mathbf{L}$ guarantees positivity, hence $\mathbf{L}\mathbf{L}^\top \in \mathcal{S}^{4}_{++}$. Furthermore, denoting $\mathbf{L}=(L_{ij})_{i,j\in \{1,2,3,4\}}$, $L_{44}=e^{-(t_x+t_y+t_z)}$ ensures $\prod_i L_{ii}=1$, so $\det(\mathbf{\Sigma}_L)=\det(\mathbf L)^2=1$. This normalizes the geometric mean of the SPD eigenvalues to $1$ without reducing the expressivity of the pose.

\subsection{Training Strategy}
\label{subsec:training}
 
 From $\mathbf L$, we decode $\hat{\mathbf t}=\big(\log L_{11},\,\log L_{22},\,\log L_{33}\big)^\top$, $\hat{\mathbf u}~=~(L_{21},L_{31},L_{41})^\top$, and $\hat{\mathbf v}=(L_{32},L_{42},L_{43})^\top$ and map $(\hat{\mathbf u},\hat{\mathbf v})$ to $\mathbf{\hat R}\in\mathcal {SO}(3)$ with the differentiable Gram–Schmidt procedure~\cite{zhou2019continuity} and supervise the training with the geodesic distance on $\mathcal {SO}(3)$ and $\ell_2$ on $\mathbb R^3$ for translation. In addition, we add two regularizers, (i) an orthogonality penalty that keeps $\langle \hat{\mathbf u},\hat{\mathbf v}\rangle\!\to\!0$ and (ii) a unit-norm penalty that prevents collapse of the two vectors. The total training loss $\mathcal{L}_{\text{pose}}$ for regressing the pose of the object ground truth pose represented by the rotation matrix $\mathbf R_{\text{gt}}$ and the translation vector $\mathbf t_{\text{gt}}$ can be expressed by,
\vspace{-3mm}

% \resizebox{0.47\textwidth}{!}{
% \begin{equation}
% \label{eq:mainloss}
%     \begin{aligned}
%     \mathcal{L}_{\text{pose}}&(\hat{\mathbf R},\hat{\mathbf t},\hat{\mathbf u},\hat{\mathbf v})
%     = \arccos\!\Big(\tfrac{\operatorname{tr}(\hat{\mathbf R}^\top \mathbf R_{\text{gt}})-1}{2}\Big)
%       + \lVert \hat{\mathbf t}-\mathbf t_{\text{gt}}\rVert_2 \\
%     &\quad + \lambda\!\left[\langle \hat{\mathbf u},\hat{\mathbf v}\rangle^2
%      + \big(\lVert \hat{\mathbf u}\rVert-1\big)^2 + \big(\lVert \hat{\mathbf v}\rVert-1\big)^2\right],
%     \end{aligned}
% \end{equation}
% }

\begin{equation}
\vspace{-1mm}
\label{eq:mainloss}
\resizebox{0.40\textwidth}{!}{%
    $
    \begin{aligned}
    \mathcal{L}_{\text{pose}}&(\hat{\mathbf R},\hat{\mathbf t},\hat{\mathbf u},\hat{\mathbf v})
    = \arccos\!\Big(\tfrac{\operatorname{tr}(\hat{\mathbf R}^\top \mathbf R_{\text{gt}})-1}{2}\Big)
      + \lVert \hat{\mathbf t}-\mathbf t_{\text{gt}}\rVert_2 \\
    &\quad + \lambda\!\left[\langle \hat{\mathbf u},\hat{\mathbf v}\rangle^2
      + \big(\lVert \hat{\mathbf u}\rVert-1\big)^2 + \big(\lVert \hat{\mathbf v}\rVert-1\big)^2\right]
    \end{aligned}
    $
}
\end{equation}

\noindent where $\lambda$ is a factor that controls the orthonormality regularization term and $\operatorname{tr}$ indicates the trace. In our experiments, the parameter $\lambda$  is set to $10^{-3}$, which stabilizes the training without dominating the main pose loss.%; thereby leading to the best results. 

\noindent For a comprehensive comparison, we also evaluate a second variant of the proposed framework with $n=3$ in which the Cholesky layer outputs a $3{\times}3$ lower–triangular matrix with $\tfrac{3(3+1)}{2}~=~6$ free entries. In this case, the rotation is parameterized by Euler angles, which are non-continuous. Further details are provided in the Supplementary.

\section{Experiments}
\label{sec:experiments}

\begin{table*}[h]
    \centering
    \setlength{\tabcolsep}{3.5pt}
    \renewcommand{\arraystretch}{0.9}
    \small
    \caption{\textbf{LM} results in terms of ADD(-S). Best result in \textbf{bold}; second best \underline{underlined}. \methodname{} (Ours) outperforms all non-P$n$P methods and slightly surpasses P$n$P-based ones.``\na'' indicates values not reported in the original paper. Classes in \textit{italics} are symmetric.} 
    \label{tab:linemod}
    \begin{tabular}{c | c  c c c c c c c c c c c c c c c}
    \toprule
    \multicolumn{2}{c}{Method} & Ape & Bvise. & Cam & Can & Cat & Drill & Duck & \textit{Eggbox} & \textit{Glue} & Holep. & Iron & Lamp & Phone & \multicolumn{1}{c}{\textbf{Avg.}} $\uparrow$ \\
    \midrule
    \multirow{3}{*}{\rotatebox{90}{Indirect}}
            & \multicolumn{1}{l|}{PVNet~\cite{peng2019pvnet}}    & 43.62 & \textbf{99.90} & 86.86 & 95.47 & 79.34 & 96.43 & 52.58 & \underline{99.15} & 95.66 & 81.92 & 98.88 & \underline{99.33} & \multicolumn{1}{l|}{92.41} & 86.27 \\
            & \multicolumn{1}{l|}{CheckerPose~\cite{lian2023checkerpose}}  & \na & \na & \na & \na & \na & \na & \na & \na & \na & \na & \na & \na & \multicolumn{1}{l|}{\na} & \underline{97.1} \\
            & \multicolumn{1}{l|}{VAPO~\cite{lian2025vapo}}         & \na & \na & \na & \na & \na & \na & \na & \na & \na & \na & \na & \na & \multicolumn{1}{l|}{\na} & \underline{97.1} \\
           
    \midrule
    \multirow{6}{*}{\rotatebox{90}{End-to-End}}
            
            & \multicolumn{1}{l|}{Pix2Pose~\cite{park2019pix2pose}}       & 58.1 & 91.0 & 60.9 & 84.4 & 65.0 & 76.3 & 43.8 & 96.8 & 79.4 & 74.8 & 83.4 & 82.0 & \multicolumn{1}{l|}{45.0} & 72.4 \\
            & \multicolumn{1}{l|}{Self6D++~\cite{wang2021occlusion}}  & 76.0 & 91.6 & \textbf{97.1} & \textbf{99.8} & 85.6 & 98.8 & 56.5 & 91.0 & 92.2 & 35.4 & \textbf{99.5} & 97.4 & \multicolumn{1}{l|}{91.8} & 85.6 \\
            & \multicolumn{1}{l|}{DeepIM~\cite{li2018deepim}}       & \underline{77.0} & 97.5 & 93.5 & 96.5 & 82.1 & 95.0 & 77.7 & 97.1 & \underline{99.4} & 52.8 & 98.3 & 97.5 & \multicolumn{1}{l|}{87.7} & 88.6 \\
            
        & \multicolumn{1}{l|}{BPnP~\cite{chen2020end}$^\dagger$}  & 74.75 & \underline{99.00} & \underline{96.25} & 98.00 & \underline{94.25} & \textbf{99.25} & \underline{78.50} & 96.50 & 90.00 & \underline{91.50} & 97.75 & \textbf{99.75} & \multicolumn{1}{l|}{\underline{97.00}} & 93.27 \\
            & \multicolumn{1}{l|}{EPro-PnP~\cite{chen2022epro}$^\dagger$}  & \na & \na & \na & \na & \na & \na & \na & \na & \na & \na & \na & \na & \multicolumn{1}{l|}{\na} & 95.80 \\
            & \multicolumn{1}{l|}{\textbf{Ours}}        & \textbf{94.8} & \textbf{99.9} & 86.0 & \underline{99.2} & \textbf{99.2} & \underline{98.9} & \textbf{89.1} & \textbf{100} & \textbf{100} & \textbf{99.5} & \underline{99.2} & 98.2 & \multicolumn{1}{l|}{\textbf{100}} & \textbf{97.2} \\
    \bottomrule
    \end{tabular}
    \captionsetup{font=small}
    \caption*{\raggedright \textsuperscript{$\dagger$} Method is End-to-End by using a differentiable P$n$P strategy.}
    \vspace{-7mm}
    
\end{table*}

In this section, we first discuss our experimental setup. Then, we report the performances of \methodname{} on three widely-used benchmarks for object pose estimation, namely, LM~\cite{10.1007/978-3-642-37331-2_42}, LM-O~\cite{10.1007/978-3-319-10605-2_35} and YCB-V~\cite{xiang2018posecnn}. We follow with a series of ablation experiments that further justify the technical choices of our solution.

\subsection{Datasets}
LineMod (LM)~\cite{10.1007/978-3-642-37331-2_42} is a well-known general pose estimation benchmark. It contains $16\text k$ images belonging to $13$ different object categories in heavily cluttered scenes. We train and test following the default BOP split and standard procedure~\cite{8803821, wang2021gdr} and thus employ $85\%$ of the RGB images for testing. LineMod Occlusion (LM-O)~\cite{10.1007/978-3-319-10605-2_35} has been introduced as a more challenging version of LM. It contains $1214$ additional annotated images and $8$ object categories under heavy occlusion. Following~\cite{wang2021gdr}, we use the publicly available physically-based rendering (pbr) images~\cite{hodavn2020bop} as training images for LM-O. Finally, the YCB-V dataset~\cite{xiang2018posecnn} is a large-scale dataset of 21 object categories and approximately $100 \text k$ real images. We follow the training strategies of~\cite{wang2021gdr} and use both the training images of YCB-V, but also the publicly available pbr images~\cite{hodavn2020bop}.

\subsection{Metrics}
We report our results using the Average Distance of Model Points (ADD) metric~\cite{10.1007/978-3-642-37331-2_42, 10.1007/978-3-319-49409-8_52} and, for symmetric objects, its symmetric variant ADD-S~\cite{xiang2018posecnn}. We compute the mean distance between the model's points transformed using the ground-truth pose and the predicted pose. If the mean distance is less than $10\%$ of the diameter of the model, the predicted pose is considered correct. The metric is then the proportion of correct poses. On the YCB-V dataset, additional results are presented in terms of AUC (Area Under the Curve) of ADD-S and AUC of ADD(-S). For computing AUC, we set the maximum distance threshold at $10$ cm~\cite{xiang2018posecnn}.

\subsection{Implementation \& Training Details}
\methodname{} is trained end-to-end with a mixed-geometry optimizer to respect manifold constraints. We also define an optimizer with separate groups to partition trainable parameters into (i) Stiefel-constrained BiMap weights $\{\mathbf W_l\}$ from~\eqref{eq:spd_head_stack} and (ii) Euclidean parameters for the backbone. The Stiefel parameters are updated with a Riemannian step, which consists of a gradient projection and QR retraction~\cite{huang2017riemannian} using an initial learning rate $10^{-2}$, while the Euclidean parameters use Adam~\cite{kingma2014adam} with an initial learning rate $10^{-4}$. We use a \texttt{ReduceLROnPlateau} scheduler~\cite{paszke2019pytorch} that halves the learning rates after $4$ epochs without improvement on the validation loss. For feature extraction, we use an ImageNet-pretrained~\cite{deng2009imagenet,tan2019efficientnet} EfficientNet-B6 backbone~\cite{tan2019efficientnet} from the Pytorch library. Its output resolution is $H{=}W{=}17$, resulting in a $289{\times}289$ spatial covariance matrix that encodes co-activations and feeds the pose head. The SPD head is constructed by $4$ BiMap interleaved with $4$ ReEig layers. The number of parameters of the model is $41.4$M. We use $4$ NVIDIA A100 GPUs (40\,GB) with batch size $8$ to train for $30$ epochs on LM and LM-O, and for $20$ epochs on YCB-V. In all datasets, $90\%$ of the training data is used for training and $10\%$ for validation. Testing is done with the best checkpoint on validation set.

\subsection{Comparison with State-of-the-art Methods}

\noindent \textbf{Results on LM.}
Table~\ref{tab:linemod} summarizes LM results. End-to-end methods predict pose directly from the input image. Indirect methods first estimate 2D–3D correspondences and then apply a P$n$P/RANSAC step. \methodname{} surpasses all end-to-end baselines, including DeepIM~\cite{li2018deepim}, and exceeds P$n$P-based methods by $0.1$ ADD(-S).

\noindent \textbf{Results on LM-O.}
As shown in Table~\ref{tab:linemod-occ}, we compare our approach with representative end-to-end methods trained per object and also include recent correspondence-based pose estimation techniques. \methodname{} outperforms all end-to-end pose learning baselines and closes the gap to leading P$n$P-based approaches such as ZebraPose~\cite{su2022zebrapose} under the same occlusion setup. %, indicating the relevance of the proposed method.

\begin{table}[h]
    \centering
    \scriptsize
    \setlength{\tabcolsep}{2pt}
    \renewcommand{\arraystretch}{1}
    \caption{\textbf{LM-O} results in terms of ADD(-S). Group-wise best in bold, group-wise second underlined. \methodname{} (Ours) outperforms all No-PnP methods and is slightly behind ZebraPose~\cite{su2022zebrapose} with only a $0.1\%$ difference.}
    \label{tab:linemod-occ}
    \resizebox{\columnwidth}{!}{%
    \begin{tabular}{c | c  c c c c c c c c c c}
    \toprule
    \multicolumn{2}{c}{Method} & Ape & Can & Cat & Drill & Duck & \textit{Eggb.} & \textit{Glu.} & Holep. & \multicolumn{1}{c}{\textbf{Avg.}} $\uparrow$ \\
    \midrule
    \multirow{4}{*}{\rotatebox{90}{Indirect}}
            & \multicolumn{1}{l|}{ZebraPose~\cite{su2022zebrapose}}
              & 57.9 & 95.0 & 60.6 & \second{94.8} & 64.5 & \second{70.9} & \second{88.7} & \multicolumn{1}{l|}{83.0} & 76.9 \\
            & \multicolumn{1}{l|}{CheckerPose~\cite{lian2023checkerpose}}
              & \second{58.3} & \second{95.7} & \underline{62.3} & 93.7 & \textbf{69.9} & 70.0 & 86.4 & \multicolumn{1}{l|}{\second{83.8}} & 77.5 \\
            & \multicolumn{1}{l|}{VAPO~\cite{lian2025vapo}}
              & \na & \na & \na & \na & \na & \na & \na & \multicolumn{1}{l|}{\na} & \second{78.02} \\
            & \multicolumn{1}{l|}{6D-Diff~\cite{xu20246d}}
              & \best{60.6} & \textbf{97.9} & \textbf{63.2} & \textbf{96.6} & \underline{67.2} & \best{73.5} & \best{92.0} & \multicolumn{1}{l|}{\textbf{85.5}} & \textbf{79.6} \\
    \addlinespace[2pt]
    \specialrule{.08em}{0pt}{0pt}
    \specialrule{.08em}{0pt}{0pt}
    \addlinespace[2pt]
    \multirow{7}{*}{\rotatebox{90}{End-to-End}}
            & \multicolumn{1}{l|}{PoseCNN~\cite{xiang2018posecnn}}
              & 9.6 & 45.2 & 0.93 & 41.4 & 19.6 & 22.0 & 38.5 & \multicolumn{1}{l|}{22.1} & 24.9 \\
            & \multicolumn{1}{l|}{Pix2Pose~\cite{park2019pix2pose}}
              & \na & \na & \na & \na & \na & \na & \na & \multicolumn{1}{l|}{\na} & 36.3 \\
            & \multicolumn{1}{l|}{Single-Stage~\cite{hu2020single}$^\dagger$}
              & 19.2 & 65.1 & 18.9 & 69.0 & 25.3 & 52.0 & 51.4 & \multicolumn{1}{l|}{45.6} & 43.3 \\
            & \multicolumn{1}{l|}{DeepIM~\cite{li2018deepim}}
              & \second{59.2} & 63.5 & 26.2 & 55.6 & \second{52.4} & \best{63.0} & 71.7 & \multicolumn{1}{l|}{52.5} & 55.5 \\
             & \multicolumn{1}{l|}{Self6D++~\cite{wang2021occlusion}}
              & 57.7 & \second{95.0} & \second{52.6} & 90.5 & 26.7 & 45.0 & \second{87.1} & \multicolumn{1}{l|}{23.5} & 59.8 \\
            & \multicolumn{1}{l|}{GDR-Net~\cite{wang2021gdr}$^\dagger$}
              & 46.8 & 90.8 & 40.5 & \second{82.6} & 46.9 & 54.2 & 75.8 & \multicolumn{1}{l|}{\second{60.1}} & \second{62.2} \\
            % & \multicolumn{1}{c|}{GDRNPP~\cite{10937121}}
              % & \na & \na & \na & \na & \na & \na & \na & \multicolumn{1}{l|}{\na} & \second{71.3} \\
            & \multicolumn{1}{l|}{\textbf{Ours}}
              & \textbf{64.0} & \best{96.4} & \textbf{74.1} & \best{96.3} & \best{55.8} & \second{57.9} & \textbf{93.6} & \multicolumn{1}{l|}{\best{76.7}} & \best{76.8} \\
    \bottomrule
    \end{tabular}%
    }
\end{table}

\noindent \textbf{Results on YCB-V.}
Table~\ref{tab:ycbv} summarizes ADD(-S) and AUC averaged over all  YCB-V objects. \methodname{} attains the best ADD(-S) among end-to-end methods and the second-best AUC after GDR-Net~\cite{wang2021gdr}, trailing by 1.6 AUC(ADD-S). P$n$P-based methods still lead overall, but \methodname{} reduces the gap \wrt the best P$n$P-based method, reducing it by 2.3 AUC(ADD-S) and 5.7 AUC(ADD(-S)).

\begin{table}[h]
\centering
\scriptsize
\setlength{\tabcolsep}{2.2pt}
\renewcommand{\arraystretch}{0.8}
\caption{\textbf{YCB-V} results in terms of ADD(-S), AUC of ADD(-S)/ADD(-S). The higher the better. Group-wise best in bold, group-wise second underlined.}
\label{tab:ycbv}
\resizebox{0.47\textwidth}{!}{%
\begin{tabular}{c | l c c c}
\toprule
\multicolumn{2}{c}{Method} & ADD(-S) & AUC of ADD-S & AUC of ADD(-S) \\
\midrule
\multirow{5}{*}{\rotatebox{90}{Indirect}}
 & \multicolumn{1}{l|}{PVNet~\cite{peng2019pvnet}}       & --    & --    & 73.4 \\
 & \multicolumn{1}{l|}{ZebraPose~\cite{su2022zebrapose}} & 80.5 & 90.1 & 85.3 \\
 & \multicolumn{1}{l|}{CheckerPose~\cite{lian2023checkerpose}} & 81.4 & 91.3 & 86.4 \\
 & \multicolumn{1}{l|}{VAPO~\cite{lian2025vapo}}  & \textbf{84.9} & \textbf{92.3} & \textbf{87.9} \\
 & \multicolumn{1}{l|}{6D-Diff~\cite{xu20246d}}   & \second{83.8} & \second{91.5} & \second{87.0} \\
\addlinespace[2pt]
\specialrule{.08em}{0pt}{0pt}
\specialrule{.08em}{0pt}{0pt}
\addlinespace[2pt]
\multirow{6}{*}{\rotatebox{90}{End-to-End}}
 & \multicolumn{1}{l|}{Pix2Pose~\cite{park2019pix2pose} }                & 45.7 & \na  & \na \\
 & \multicolumn{1}{l|}{Single-Stage~\cite{hu2020single}$^\dagger$}       & 53.9  & \na   & \na \\
 & \multicolumn{1}{l|}{PoseCNN~\cite{xiang2018posecnn}}  & \na  & 75.9 & 61.3 \\
 & \multicolumn{1}{l|}{DeepIM~\cite{li2018deepim}}       & \na  & 88.1 & 81.9 \\
 & \multicolumn{1}{l|}{GDR-Net~\cite{wang2021gdr}$^\dagger$}       & \second{60.1} & \textbf{91.6} & \textbf{84.4}\\
 % & \multicolumn{1}{c|}{GDRNPP~\cite{10937121}}          & \best{82.5} & \na  & \na \\
 & \multicolumn{1}{l|}{\textbf{Ours}}                             & \textbf{69.7} & \second{90.0} & \second{82.2} \\
\bottomrule
\end{tabular}%
}
\end{table}

\subsection{Ablation Study}
\label{sec:ablation}
We conduct ablations on the LM-O dataset and report performance averaged over all objects in terms of ADD(-S).

\noindent \textbf{Impact of the Rotation Representation.}
In our framework, the predicted pose is embedded in a $4{\times}4$ Cholesky factor whose off–diagonal entries carry a 6D rotation parameterization $(\mathbf u,\mathbf v)\in\mathbb{R}^{3\times2}$. The latter is mapped to $\mathcal{SO}(3)$ via a differentiable Gram–Schmidt step, which produces a continuous training signal on the rotation manifold $\mathcal{SO}(3)$. 
As discussed at the end of Section~\ref{sec:methodology}, we also evaluate a lower–dimensional variant that encodes rotation with Euler angles. 
In this setting, we regress ${\mathbf \Sigma}_L\in\mathcal{S}^{3}_{++}$ and decode the pose from its $3{\times}3$ Cholesky factor. 
Further details of this configuration are in the supplementary. Table~\ref{tab:rot-abl} shows that the 6D representation leads to a higher ADD(-S) than Euler angles, reflecting the benefit of continuity.

\begin{table}[!h]
\centering
\caption{Comparison with an alternative rotation parameterization. Under identical settings, we replace our \textbf{6D + differentiable GS} rotation with an Euler-angle parameterization encoded as a $3{\times}3$ SPD. The 6D+GS choice results in higher performance.}
\label{tab:rot-abl}
\footnotesize
\setlength{\tabcolsep}{3pt}
\renewcommand{\arraystretch}{0.}
\resizebox{\linewidth}{!}{%
\begin{tabular}{@{}lccc@{}}
\toprule
Rep. & SPD size & Cont.\ on $\mathcal{SO}(3)$ & Avg.\ ADD(-S)$\uparrow$ \\
\midrule
Euler angles        & $3{\times}3$ & \xmark & 70.9 \\
\textbf{6D+GS (Ours)} & $4{\times}4$ & \cmark & \textbf{76.8} \\
\bottomrule
\end{tabular}}
\vspace{-3mm}
\end{table}

\noindent\textbf{Impact of the SPD Head Components.}
We ablate the manifold-aware head and the structured pose encoding we are using to isolate the effect of each. To investigate this, we design three additional variants.
(A) We replace the SPD head with 2 FC layers that map a vectorized version of ${\boldsymbol {\hat \Sigma}}$ directly to the pose. This assesses the benefit of the SPD layers in learning the compact SPD matrix that is fed to the Cholesky layer. (B) We keep the CovPooling layer, but use it to compute \emph{channel} covariances instead of \emph{spatial} covariance and feed $\mathbf{\hat \Sigma}_c$ to the same SPD head. This contrasts the spatial covariance to the channel covariance. (C) We keep the SPD head and drop the Cholesky layer and train using the Frobenius norm between the projection of $\mathbf{\Sigma}_L$ to the logarithmic tangent space and the precomputed corresponding SPD pose label. This assesses the effect of our pose decoding to $(\mathbf R,\mathbf t)$ and the calculation of the training loss on $\mathcal SO(3){\times}\mathbb R^3$ instead of on the log-tangent space. All other settings are kept fixed. Table~\ref{tab:ablation_spd} shows that our full model attains the best performance. Indeed, the replacement of the SPD head with a Euclidean MLP head causes a sharp drop in accuracy, reflecting a mismatch between the Euclidean hypothesis and the SPD geometry of the representation.

\begin{table}[h]
\caption{Comparison between \textbf{Ours}, (A) Euclidean regression head, (B) Channel-wise covariance, and (C) Training on log tangent space of SPD manifold.}
\label{tab:ablation_spd}
\centering
\setlength{\tabcolsep}{5pt}
\renewcommand{\arraystretch}{0.9}
\begin{tabular}{lcccc}
\toprule
Variant & (A) & (B) & (C) & \methodname{} \\
\midrule
Avg.\ ADD(-S)$\uparrow$ & 31.0 & 70.9 & 72.3 & \textbf{76.8} \\
\bottomrule
\end{tabular}
\vspace{-4mm}
\end{table}

\begin{figure}[h]
    \centering
    \includegraphics[width=0.4\textwidth]{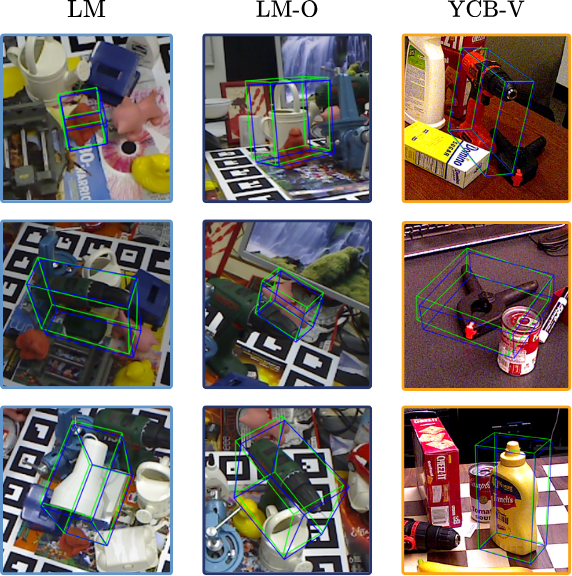}
    \caption{\textbf{Qualitative results.} \textcolor{MyGreen}{Green} 3D boxes denote GT poses, whereas \textcolor{MyBlue}{blue} 3D boxes denote predictions. \methodname{} predicts accurate poses in cluttered scenes and under partial occlusion.}
    \label{fig:qualitative}
\end{figure}

\subsection{Inference Time}
Inference time is measured on a machine with an AMD $3.30$GHz CPU with an NVIDIA A100 GPU. The backbone takes $22.6$ms for feature extraction, the covariance-pooling layer $0.5$ms to compute the covariance matrix, and the head $23.8$ms to produce the compact $4{\times}4$ representation and regress the pose, for a total of $46.9$ms. Table~\ref{tab:method-times} reports inference time for running state-of-the-art methods. \methodname{} achieves the best trade-off between accuracy and speed compared to previous approaches.

\begin{table}[h]
\centering
\begingroup
\caption{Comparison of inference time and ADD-(S) across datasets. We measured inference time of the open-source baselines on the same hardware as ours.}
\label{tab:method-times}
\small % slightly smaller than \small; switch to \scriptsize if needed
\setlength{\tabcolsep}{1.0pt}
\renewcommand{\arraystretch}{0.8}
\begin{tabularx}{0.47\textwidth}{c | X c c @{\hspace{1pt}} c @{\hspace{1pt}} c}
\toprule

\multicolumn{2}{c}{\multirow{2}{*}{Method}} &
\multicolumn{1}{c}{Time} &
\multicolumn{3}{c}{ADD-(S)} \\
\multicolumn{1}{c}{}&& (ms) &
LM~\cite{10.1007/978-3-642-37331-2_42} &
LM-O~\cite{10.1007/978-3-319-10605-2_35} &
YCB-V~\cite{xiang2018posecnn} \\

\midrule
\multirow{4}{*}{\rotatebox[origin=c]{90}{\scriptsize Indirect}}
& \multicolumn{1}{l|}{PVNet~\cite{peng2019pvnet}}          & 44.9    & 86.27 & \na   & \na   \\
& \multicolumn{1}{l|}{ZebraPose~\cite{su2022zebrapose}}    & 119.3   & \na   & 76.9  & 80.5  \\
& \multicolumn{1}{l|}{CheckerPose~\cite{lian2023checkerpose}} & 116.7 & 97.1  & 77.5  & 81.4  \\
& \multicolumn{1}{l|}{VAPO~\cite{lian2025vapo}}            & 69.8  & 97.1  & 78.02 & 84.9  \\
\midrule
\multirow{4}{*}{\rotatebox[origin=c]{90}{\scriptsize End-to-End}}
& \multicolumn{1}{l|}{Pix2Pose~\cite{park2019pix2pose}}    & 123   & 72.4  & 36.3  & 45.7  \\
& \multicolumn{1}{l|}{DeepIM~\cite{li2018deepim}}          & 77.3    & 88.6  & 55.5  & \na   \\
& \multicolumn{1}{l|}{GDR\text{-}Net~\cite{wang2021gdr}}   & 16    & \na   & 62.2  & 60.1  \\
& \multicolumn{1}{l|}{\textbf{Ours}}                       & \textbf{46.9} & \textbf{97.2} & \textbf{76.8} & \textbf{69.7} \\
\bottomrule
\end{tabularx}
\endgroup
\vspace{-3mm}
\end{table}

\section{Conclusion \& Future Directions}
\label{sec:conclusion}

We present \methodname, an end-to-end framework for direct 6-DoF object pose estimation from a single RGB image. \methodname{} performs spatial covariance pooling and learns directly on the symmetric positive-definite (SPD) manifold to regress object pose. Concretely, we encode the image as a spatial covariance matrix that captures co-variation among image regions, then compress it into a compact SPD code via a geometry-aware, end-to-end–trainable set of layers. A differentiable Cholesky decomposition acting on the SPD network output results in a continuous 6D rotation representation and a translation recovering the pose. Across 3 BOP benchmarks, \methodname{} attains state-of-the-art accuracy for end-to-end single-object pose estimation and substantially reduces the gap as compared to correspondence-based approaches. 
As future directions, we will investigate symmetry-aware training and inference for pose-ambiguous objects within a CAD-free setup, while preserving the fully differentiable end-to-end nature of \methodname{}. Currently, \methodname{} does not explicitly handle object symmetries, and addressing this limitation could be an interesting direction for future work. A symmetry-focused pilot study and discussion are provided in the supplementary.

\noindent \textbf{Acknowledgement.} The present work is supported by the National Research Fund (FNR), Luxembourg, under the C21/IS/15965298/ELITE project, and by Infinite Orbits. Experiments were performed on the Luxembourg national supercomputer MeluXina at LuxProvide.

 {
    \small
    \bibliographystyle{ieeenat_fullname}
    \bibliography{main}
}

\clearpage
\maketitlesupplementary

\setcounter{section}{0}
\setcounter{equation}{0}
\setcounter{figure}{0}
\setcounter{footnote}{0}
\setcounter{table}{0}

\renewcommand\thesection{\Alph{section}}
\renewcommand{\theequation}{\alph{equation}}
\renewcommand{\thefigure}{S\arabic{figure}}
\renewcommand{\thetable}{S\arabic{table}}
\renewcommand{\thefootnote}{\fnsymbol{footnote}}

% Supplementary Abstract
This supplementary material presents implementation details, detailed ablation studies, a pilot discussion on symmetry-aware training and extended experiments for \methodname{}. We describe optimization and architecture choices, ablate key components of the proposed method, and report detailed per-object and cross-dataset results, as well as additional validation on Absolute camera Pose Regression (APR) on Cambridge Landmarks~\cite{kendall2015posenet} and the more challenging spacecraft pose estimation task on SPEED+~\cite{park2022speed+}, further demonstrating the applicability of the proposed method.

\section{Additional Implementation Details}
\label{supp:training}

\noindent\textbf{Ground-truth Preprocessing.} As shown in~(\textcolor{cvprblue}{9}) the 6D rotation and the 3D translation are encoded in the non-zero elements of the lower-triangular matrix $\mathbf L$. For rotation, we use the full $\mathbf R_{\text{gt}}\in \mathcal SO(3)$ matrix as the rotation target/objective. For translation, we compute per-axis statistics on the training split, \ie, 99$^{\text{th}}$-percentile-based per-axis minima $\mathbf t_{\min}$ and per-axis ranges $\mathbf t_{\mathrm{range}}$, and form normalized targets such that,
\begin{equation}
\mathbf t_{\text{gt}}=\big(\mathbf t_{\text{raw}}-\mathbf t_{\min}\big) / \mathbf t_{\mathrm{range}},
\end{equation}
\noindent where $\mathbf t_\text{raw}$ are the original translation values from the dataset. We train with the normalized translation as objective and at inference, the network predicts ${\mathbf t^{{'}}}$ which we de-normalize with the same fixed statistics,
\begin{equation}
\mathbf t=\mathbf t^{{'}} \times \mathbf t_{\mathrm{range}}+\mathbf t_{\min}.    
\end{equation}
\noindent Note that the division and product here are performed axis-wise. This preprocessing keeps targets well-scaled without changing the metric translation recovered at test time.

\noindent \textbf{Stiefel–constrained Optimizer for BiMap Weights.} We maintain each projection matrix at layer $l$ on the Stiefel manifold $\mathbf W_l \in \mathcal{V}_n(\mathbb{R}^m)$ using a projected gradient step
followed by a QR–based retraction. Given a gradient $\mathbf G~=~\nabla_{\mathbf W}\mathcal{L}_{\text{pose}}$, where $\mathcal{L}_{\text{pose}}$ is the training loss defined in~(\textcolor{cvprblue}{10}), we first project this gradient onto the tangent space at $\mathbf W_l$,
\begin{equation}
\mathbf G_{\!\top} \;=\; \mathbf G \;-\; \mathbf W_l\,\operatorname{sym}\!\big(\mathbf W^\top_{l} \mathbf G\big),
\end{equation}
where $\operatorname{sym}$ denotes the symmetrization operator\footnote{For a square matrix
$\mathbf A$,  $\operatorname{sym}(\mathbf A) = \tfrac{1}{2} (\mathbf A + \mathbf A^\top)$.}.
We then take a step along $-\mathbf G_{\!\top}$ with step size $\eta$,
\begin{equation}
\widetilde{\mathbf W}_{l} \;=\; \mathbf W_{l} \;-\; \eta\, \mathbf G_{\!\top},
\end{equation}
and retract back to $\mathcal{V}_n(\mathbb{R}^m)$ via the $\mathbf Q$ factor of a reduced QR decomposition,
\begin{equation}
\widetilde{\mathbf W}_{l} \;=\; \mathbf Q_l \mathbf R_l,
\qquad
\mathbf W^{+}_{l} \;=\; \mathbf Q_l,
\end{equation}
where $\mathbf W^{+}_{l}$ denotes the updated Stiefel-constrained weight at layer $l$ after the retraction. In our implementation, we perform the QR decomposition in double precision when enabled, for improved numerical stability.

\section{Additional Details on Ablation Studies}
\label{supp:add_ablation}
We provide additional details on the variants that we design in Section~\textcolor{cvprblue}{5.5}. Table~\ref{supp:ablation} shows the results on LM-O~\cite{10.1007/978-3-319-10605-2_35} per-object of the ablation experiments.

\begin{table}[h]
\caption{Per-object results on \textbf{LM-O}. We compare our full method with ablated variants of \methodname{} to isolate the contribution of each component of the SPD head. This table is the detailed per-object version of Table~\textcolor{cvprblue}{5}.}
\label{supp:ablation}
\centering
\setlength{\tabcolsep}{5pt}
\renewcommand{\arraystretch}{0.8}
\begin{tabular}{lcccc}
\toprule
Variant & (A) & (B) & (C) & \methodname{} \\
\midrule
Ape          & 27.2 & 56.1 & 62.6 & 64.0 \\
Can          & 40.8 & 93.8 & 94.4 & 96.4 \\
Cat          & 15.7 & 70.0 & 66.1 & 74.1 \\
Driller      & 26.7 & 88.1 & 91.9 & 96.3 \\
Duck         & 16.7 & 49.3 & 48.1 & 55.8 \\
\textit{Eggbox}      & 47.5 & 48.0 & 53.6 & 57.9 \\
\textit{Glue}        & 42.4 & 91.7 & 87.0 & 93.6 \\
Holep.    & 30.9 & 70.1 & 74.7 & 76.8 \\
\midrule
Avg.\ ADD(-S)$\uparrow$ & 31.0 & 70.9 & 72.3 & \textbf{76.8} \\
\bottomrule
\end{tabular}
\end{table}

\begin{figure}[h]
    \centering
    \includegraphics[width=\linewidth]{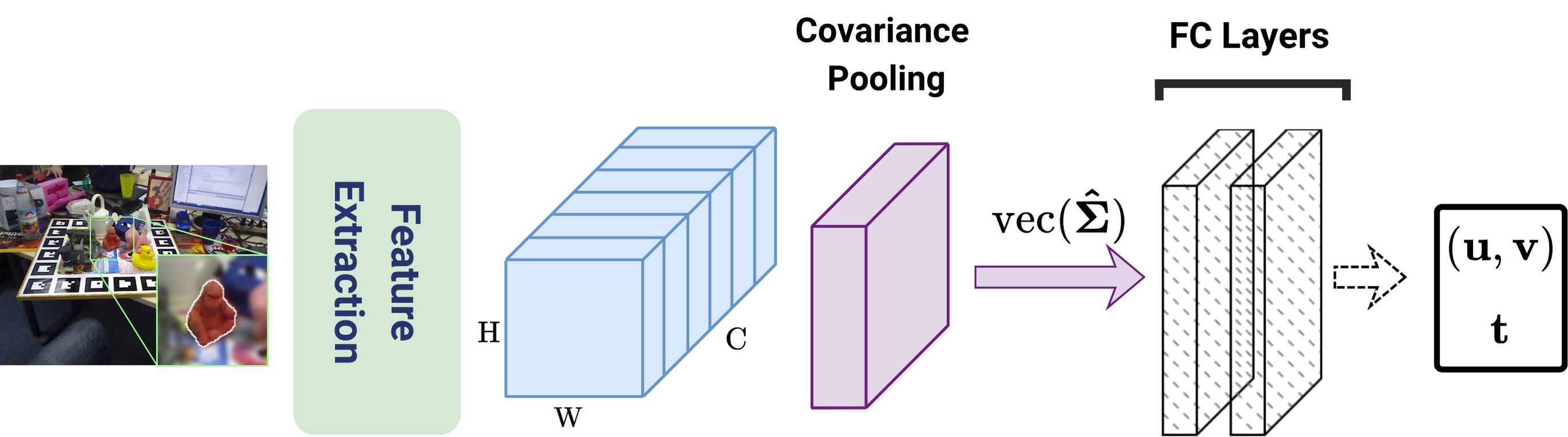}
    \caption{\textbf{Variant (A)}. The spatial covariance matrix $\boldsymbol{\hat \Sigma}$ is vectorized and fed to a two-layer MLP that regresses pose purely in Euclidean space.}
    \label{supp:fc}
\end{figure}

\noindent\textbf{(A) Euclidean Regression Head.}
We replace the regression head with a purely Euclidean baseline composed of a two–layer MLP as illustrated in Figure~\ref{supp:fc}. From spatial covariance $\boldsymbol{\hat \Sigma}\in\mathcal{S}^{N}_{++}$ we form a vector $\operatorname{vec}(\boldsymbol{\hat \Sigma})\in\mathbb{R}^{\tfrac{N(N+1)}{2}}$ by stacking its upper–triangular entries \ie unique elements. The MLP maps $\boldsymbol{\hat \Sigma} \!\to\! \mathbb{R}^{256} \!\to\! \mathbb{R}^{9}$, directly regressing the 6D rotation parameters $(\mathbf u,\mathbf v)$ and the 3D translation $\mathbf t$. As reported in Table~\textcolor{cvprblue}{5}, this setup significantly decreases performance because learning in Euclidean space ignores the Riemannian geometry of the SPD manifold and collapses the second-order spatial structure.

\noindent\textbf{(B) Channel-wise Covariance.} Given backbone features $\mathbf F\!\in\!\mathbb{R}^{C\times H\times W}$, we flatten the spatial grid to obtain $\mathbf X\!\in\!\mathbb{R}^{C\times N}$ with $N{=}HW$, where each row is a channel and each column a spatial location. We remove the spatial mean per channel to obtain $\tilde{\mathbf X}$ and form the channel-wise covariance,

\begin{equation}
\boldsymbol{\hat \Sigma}_{\text{ch}}
\;=\; \tfrac{1}{N-1}\,\tilde{\mathbf X}\,\tilde{\mathbf X}^{\top}
\;\in\;\mathcal{S}^{C}_{++}.
\end{equation}

This descriptor summarizes second-order co-activations between channels aggregated over all spatial locations. Table~\textcolor{cvprblue}{5} shows that spatial covariance empirically surpasses channel covariance, as it preserves pairwise relations between image regions that vary with viewpoint, whereas channel covariance collapses the spatial layout and cannot distinguish configurations with different spatial arrangements. It also requires more operations to construct; Forming channel covariance costs \(\mathcal{O}(C^2N)\), while the spatial covariance costs \(\mathcal{O}(N^2C)\). Since in late backbone stages \(C\gg N\) (in our setup \(C{=}2304\), \(H{\times}W{=}17{\times}17\Rightarrow N{=}289\)), channel covariance is markedly heavier. Empirically, on the same device, it runs at \(168\,\mathrm{ms}\) per image, versus \(46.9\,\mathrm{ms}\) for spatial covariance, further supporting the use of spatial covariance.

\begin{figure}[h]
    \centering
    \includegraphics[width=\linewidth]{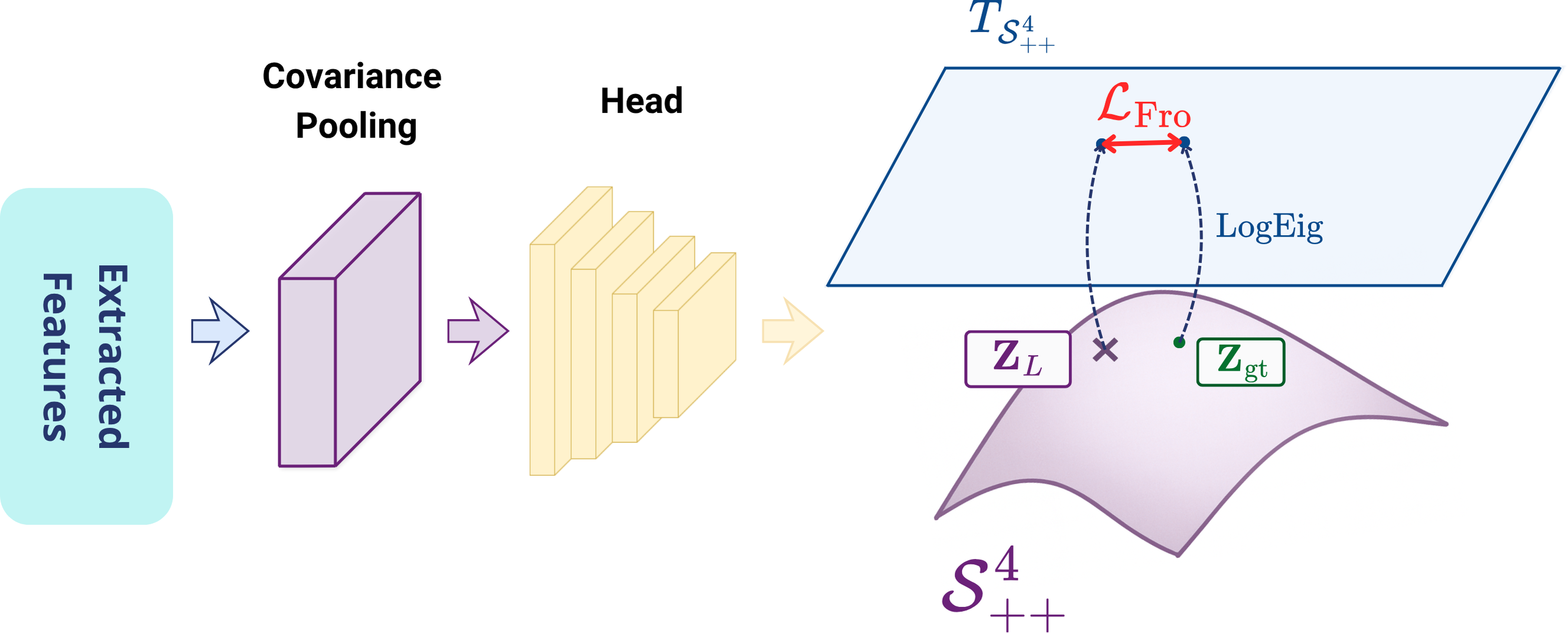}
    \caption{\textbf{Variant (C)}. We remove the Cholesky-based pose decoder and train on the log-tangent space of the SPD manifold. $\mathbf Z_L$ and $\mathbf Z_{\text{gt}}$ are mapped via $\operatorname{LogEig}$ to the log–tangent space. For clarity, a single tangent space $T_{\mathcal{S}_{++}^4}$ is depicted.}
    
    \label{supp:tangent}
\vspace{-2mm}
\end{figure}

\noindent\textbf{(C) Training on the log–tangent space of the SPD manifold.}
This variant keeps the SPD head and removes the Cholesky pose decoder. Let ${\mathbf Z}_L\in\mathcal S_{++}^{4}$ be the SPD output of the head composed of $L$ layers, and let $\mathbf Z_{\text{gt}}\in\mathcal S_{++}^{4}$ be the precomputed SPD pose label obtained from the ground-truth pose in a manner similar to Equation~(\textcolor{cvprblue}{9}). We train by aligning both in the tangent space using a Log–Euclidean map and use a Frobenius loss
\begin{equation}
\mathcal L_{\text{Fro}}=\big\|\operatorname{LogEig}\!\big(\mathbf Z_L\big)-\operatorname{LogEig}\!\big(\mathbf Z_{\text{gt}}\big)\big\|_F^2 \ ,
\end{equation}

\noindent where $\operatorname{LogEig}$ is the matrix logarithm operation~\cite{huang2017riemannian}. No pose loss is used here; this design shows the relevance of using our Cholesky decoder to learn the pose parameters. The process is illustrated in Figure~\ref{supp:tangent}.

\section{Rotation Representation using Euler Angles}
\label{supp:3x3}

To assess the benefit of using a continuous rotation representation in $\mathcal{SO}(3)$, we also design an alternative encoding using a lower–dimensional variant that encodes the rotation using Euler angles which are known to be discontinuous due to Gimbal lock~\cite{zhou2019continuity}. In this variant, we regress ${\mathbf Z}_L\in\mathcal{S}^{3}_{++}$ and the mapping from~(\textcolor{cvprblue}{8}) becomes 
\begin{equation}
    \begin{array}{cccccc}
        \mathbf{\Psi_{-}} & : & \mathcal{S}^3_{++} & \longrightarrow & \mathcal P_{-} \\
         & & {\mathbf Z}_L & \longmapsto & \mathbf{\Psi_{-}} ({\mathbf Z}_L) = \mathbf{P} \ , 
    \end{array}
    \label{supp:psi_minus}
\end{equation}

\noindent where \(\mathbf{P} = (\boldsymbol{\theta}, \mathbf{t})\) is the pose with \(\mathbf t=(t_x,t_y,t_z) \in \mathbb{R}^3\) the translation vector and \(\boldsymbol{\theta} = (\theta_x, \theta_y, \theta_z) \in [0,2\pi)^3\) the three Euler angles (roll, pitch and yaw) that represent the object's rotation in 3D space \wrt the camera frame. The corresponding pose space is $\mathcal{P}_{-} = [0,2\pi)^3 \times \mathbb{R}^3$ which is a discontinuous pose representation due to the wrap–around of the Euler angles (at \(0/2\pi\)) and the gimbal–lock singularities. For encoding the 3D rotation representation (3-dimensional) as well as the translation vector (3-dimensional) we need a matrix with at least 3 non-zero, thus fixing $n$ to 3, the Cholesky decomposition such that ${\mathbf Z}_L= \mathbf{L}\mathbf{L}^\top\in\mathcal{S}^{3}_{++}$ gives a $3{\times}3$ lower-triangular matrix that we use to encode the pose parameters such that,

\begin{equation}
\label{supp:3x3-cholesky}
    \begin{aligned}
    \mathbf{L} &=
    \begin{bmatrix}
    e^{\theta_x} & 0 & 0\\
    t_x & e^{\theta_y} & 0 \\
    t_y & t_z & e^{\theta_z} \\
    \end{bmatrix} \ .
    \end{aligned}
\end{equation}

\noindent For training, we keep an identical configuration as the default 6D + differentiable GS we propose for our method, the results comparing the two rotation parameterizations are shown in Table~\textcolor{cvprblue}{4} of the main paper.

\begin{figure*}[ht]
    \centering
    \includegraphics[width=\textwidth]{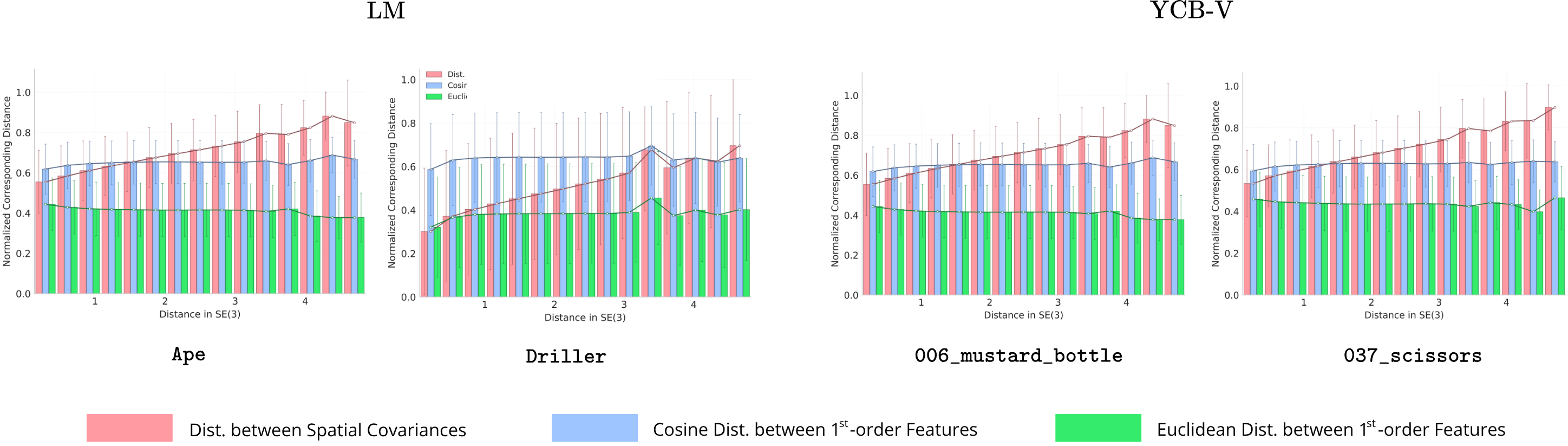}
   
    \caption{\textbf{Additional Pose–representation alignment.} As in Figure~\textcolor{cvprblue}{1}, image pairs are sorted by ground-truth $\text{SE}(3)$ distance. We plot Log–Euclidean distances between spatial covariances and, for comparison, cosine and Euclidean distances between flattened features. Covariance distances grow with distances in $\text{SE}(3)$, while flattened-feature distances remain nearly constant, reinforcing the hypothesis that spatial covariances are more correlated with true pose.}

    \label{supp:motivation2}
\end{figure*}

\section{Computation and memory requirements}
We further provide a brief analysis of the scaling trade-offs between the spatial covariance size and the spatial resolution ($N{\times}N$). As discussed in Section~\ref{supp:add_ablation}, CovPool has time complexity $\mathcal{O}(N^2 C)$ and memory complexity $\mathcal{O}(N^2)$. Table~\ref{supp:computation} shows that the CovPool runtime is negligible (less than 1\,ms), while the SPD head dominates the computation and increases with $N$. Therefore, we use, and recommend using, late, low-resolution features, or keeping the spatial resolution moderate via downsampling/pooling. 

\begin{table}[h]
\centering
\small
\caption{CovPool layer and SPD head latencies across different spatial resolutions.}
\label{supp:computation}

\begin{tabular}{c c c}
\hline
\multicolumn{1}{c}{\multirow{2}{*}{$H{\times}W$}} & \multirow{1}{*}{CovPool} & \multirow{1}{*}{SPD Head} \\
 & (ms) &(ms) \\
\hline
$8^2$  & 0.38 & 7.8  \\
$\mathbf{17^2}$ & \textbf{0.50} & \textbf{23.8} \\
$25^2$ & 0.61 & 68.6 \\
\hline
\end{tabular}%

\end{table}

Regarding memory usage, the same setup described in Section~\ref{supp:add_ablation} requires $\approx$2.54\,MB to store $\mathbf{X}$ and $\approx$0.32\,MB to store $\mathbf{\Sigma}$ per input (both in fp32). During backpropagation, activations are also retained, resulting in an overall memory scaling of $\mathcal{O}(CN + N^2)$. Additional buffers (e.g., centered $\mathbf{X}$ and eigen/QR workspaces) have the same asymptotic order.

\section{Generality beyond CNN backbones}
We show that~\methodname{} also extends beyond CNN backbones. To illustrate this, we use a ViT\_b\_16 backbone~\cite{dosovitskiy2020image}. Instead of computing the covariance over spatial locations (pixels), we compute it over the patch tokens extracted by the ViT backbone. Importantly, because ViTs use global self-attention, the resulting SPD matrix does not represent purely ``spatial'' covariance. Nevertheless, a small pilot experiment with ViT\_b\_16 indicates that a token-covariance + SPD head still outperforms a ViT\_b\_16 baseline that regresses pose parameters using an MLP. The results are shown in Table~\ref{supp:vit_cov2pose}.

\begin{table}[h]
    \centering
    \caption{Pilot comparison on \textit{Ape} (LM-O) with a ViT\_b\_16 backbone. Results reported in ADD.}
    \label{supp:vit_cov2pose}
    \begin{tabular}{l c}
        \hline
        Class (Ape; LM-O) & ADD$\uparrow$ \\
        \hline
        ViT+MLP & 33.3 \\
        \textbf{ViT+Cov+SPD head} & \textbf{40.6} \\
        \hline
    \end{tabular}%
\vspace{-5mm}
\end{table}

\section{Pilot study on symmetry handling and point-based losses}
\label{sec:supp_symmetry_point_loss}

Object symmetries remain a well-known challenge in 6D object pose estimation. \methodname{} does not explicitly model symmetries, as its main design targets a CAD-free, RGB-only, geometry-agnostic, and fully differentiable formulation. Nevertheless, in the main results we did not observe a specific degradation on the symmetric LM/LM-O objects \textit{eggbox} and \textit{glue}, motivating a small pilot study to assess whether adding symmetry-oriented supervision improves performance.

A natural way to address symmetric ambiguities is to use point-based losses such as ADD-S, or symmetry-aware rotation losses~\cite{wang2021gdr, park2019pix2pose}. However, these losses require CAD model points and/or symmetry annotations during training. This changes the training assumptions of \methodname{}.

\paragraph{Pilot experiment.}
We trained variants of \methodname{} on the symmetric objects of LM-O by replacing the pose supervision with: (i) an ADD-S-inspired point-based loss that matches predicted and ground-truth transformed 3D points by nearest neighbors, and (ii) a symmetry-aware rotation loss from prior work similar to~\cite{wang2021gdr}. We evaluate on the symmetric objects \textit{Eggbox} and \textit{Glue} using ADD-S, following the standard BOP protocol.

\paragraph{Results.}
Table~\ref{tab:sym_pilot} summarizes the pilot results. The ADD-S-inspired loss does not improve performance in this setting and can degrade results, while the symmetry-aware loss matches or only slightly improves the baseline. In particular, on \textit{Eggbox}/\textit{Glue}, where \methodname{} achieves 57.9/93.6 ADD-S, the ADD-S-inspired variant reaches 55.0/84.8, and the symmetry-aware loss reaches 57.8/93.8.

\paragraph{Discussion.}
These pilot results suggest that simply plugging in point-based or symmetry-aware losses is not sufficient to consistently improve \methodname{}, despite introducing stronger geometric supervision. Moreover, the two pilot variants depart from the original formulation by relying on operations/supervision that break the fully differentiable end-to-end design of \methodname{} and its CAD-free training assumptions. We therefore view explicit symmetry handling as an important but non-trivial extension, especially under the constraints targeted by our framework. Designing symmetry-aware training/inference strategies that preserve the CAD-free and fully differentiable nature of \methodname{}, as in~\cite{cai2022sc6d, pitteri2019object}, is a promising future direction.

\begin{table}[h]
\centering
\footnotesize % Changed from \small to \footnotesize
\caption{Pilot study on symmetric LM-O objects using point-based and symmetry-aware losses.}
\label{tab:sym_pilot}
% Removed \resizebox to prevent the text from being stretched/blown up
\setlength{\tabcolsep}{6pt} % Increased slightly for better readability now that it's not stretched
\renewcommand{\arraystretch}{1.1} % Increased slightly so text isn't cramped vertically
\begin{tabular}{l c cc}
\hline
\multirow{2}{*}{Method} & \multirow{2}{*}{CAD?} & \textit{Egg.} & \textit{Glu.} \\
 &  & ADD-S$\uparrow$ & ADD-S$\uparrow$ \\
\hline
\methodname                  & $\times$ & \textbf{57.9} & \underline{93.6} \\
\methodname{}+ $\mathcal{L}_{\text{ADD}\text{-}\text{S}}$  & \checkmark & 55.0 & 84.8 \\
\methodname{}+ $\mathcal{L}_{\mathbf R\text{,sym}}$~\cite{wang2021gdr}        & \checkmark & \underline{57.8} & \textbf{93.8} \\
\hline
\end{tabular}
\end{table}

\section{Additional Results \& Analysis for Figure~\textcolor{cvprblue}{1}}
\label{supp:fig1}

In Figure~\textcolor{cvprblue}{1}, we show that Log–Euclidean distances between spatial covariances of CNN-extracted feature maps grow with ground-truth $\text{SE}(3)$ pose distance, whereas cosine distances between flattened features remain nearly constant. This reinforces the hypothesis that a spatial–covariance representation is more pose-sensitive than flattened activations, and thus more suitable for direct regression.

To achieve the result represented in Figure~\textcolor{cvprblue}{1}, we use an EfficientNet-B6~\cite{tan2019efficientnet} backbone pretrained on ImageNet, with no pose-specific finetuning and no augmentation to avoid task leakage and isolate the representational contrast between spatial covariance and flattened features.

Figure~\ref{supp:motivation2} presents analogous experiments on additional classes from LineMod~\cite{10.1007/978-3-642-37331-2_42} and YCB-V~\cite{xiang2018posecnn} training data. We report cosine distances, and additionally Euclidean distances, between flattened features, both of which remain nearly constant across pose separations, while Log Euclidean distances between spatial covariances increase, confirming that spatial covariance is a pose-sensitive representation and more correlated with the poses.

\section{Additional Results on APR}
\label{supp:apr}

To assess generality beyond object pose, we evaluate \methodname{} on the task of Absolute camera Pose Regression (APR) using the Cambridge Landmarks benchmark~\cite{kendall2015posenet}. The dataset comprises four outdoor scenes (King’s College, Old Hospital, Shop Facade and St Mary’s Church) with labeled video frames and standard train/test splits. We keep the same pipeline using spatial covariance pooling, BiMap+ReEig SPD head, and the SPD(4) Cholesky decoder. Following the PoseNet protocol, we report median translation [m] and rotation [$^\circ$] errors. As shown in Table~\ref{supp:apr}, compared with the PoseNet family of baselines (PoseNet~\cite{kendall2015posenet}, Bayesian PoseNet~\cite{kendall2016modelling}, and GPoseNet~\cite{kendall2017geometric}), \methodname{} achieves lower median translation and rotation errors across all scenes and on average, indicating that spatial second-order cues and manifold-aware learning transfer effectively to scene-centric APR.

\begin{table}[h]
\centering
\small
\caption{APR results on \textbf{Cambridge Landmarks}~\cite{kendall2015posenet}. Median translation [m] and rotation [$^\circ$] errors per scene and their average (Lower is better). \methodname{} achieves the best performance across all scenes and on average.}
\label{supp:apr}
\resizebox{\linewidth}{!}{
\setlength{\tabcolsep}{1pt}
\begin{tabular}{l|cccc|c}
\toprule
Method & K.~College & Hospital & Shop~Facade & St.~Mary & Avg.$\downarrow$ \\
\midrule
PoseNet~\cite{kendall2015posenet} & 1.92,\,5.40$^\circ$ & 2.31,\,5.38$^\circ$ & 1.46,\,8.08$^\circ$ & 2.65,\,8.48$^\circ$ & 2.08,\,6.83$^\circ$ \\
BayesianPN~\cite{kendall2016modelling}    & 1.74,\,4.06$^\circ$ & 2.57,\,5.14$^\circ$ & 1.25,\,7.54$^\circ$ & 2.11,\,8.38$^\circ$ & 1.91,\,6.28$^\circ$ \\
GPoseNet~\cite{kendall2017geometric}     & 1.61,\,2.29$^\circ$ & 2.62,\,3.89$^\circ$ & 1.14,\,5.73$^\circ$ & 2.93,\,6.46$^\circ$ & 2.07,\,4.59$^\circ$ \\
\textbf{Ours}   & \textbf{1.57},\,\textbf{1.81}$^\circ$ & \textbf{2.08},\,\textbf{2.35}$^\circ$ & \textbf{1.11},\,\textbf{5.33}$^\circ$ & \textbf{2.06},\,\textbf{5.00}$^\circ$ & \textbf{1.70},\,\textbf{3.56}$^\circ$ \\
\bottomrule
\end{tabular}}
\end{table}

\vspace{-4mm}
\section{YCB-V Dataset Per-object Evaluation}
\label{supp:ycbv-perObject}
In Table~\ref{supp:tab-ycbv}, we report a more detailed overview of the results per-object on the YCB-V dataset~\cite{xiang2018posecnn}. For the sake of space, we report only the results of End-to-End pose learning methods.

\begin{table*}[h]
\centering
\scriptsize
\setlength{\tabcolsep}{1pt}
\caption{\textbf{YCB-V} per-object results in terms of ADD(-S) and AUC-S/AUC(-S).``\na'' indicates values not reported in the original paper. Classes denoted with \texttt{$^{*}$} are symmetric.} 
\label{supp:tab-ycbv}
\resizebox{\textwidth}{!}{%
\begin{tabular}{l|ccc|ccc|ccc|ccc|ccc}
\hline
\multicolumn{1}{c|}{\multirow{3}{*}{Method}} &
\multicolumn{3}{c|}{PoseCNN~\cite{xiang2018posecnn}} &
% \multicolumn{3}{c|}{Pix2Pose~\cite{park2019pix2pose}} &
\multicolumn{3}{c|}{Single-Stage~\cite{hu2020single}$^\dagger$} &
\multicolumn{3}{c|}{DeepIM~\cite{li2018deepim}} &
\multicolumn{3}{c|}{GDR-Net~\cite{wang2021gdr}$^\dagger$} &
\multicolumn{3}{c}{\textbf{Ours}} \\
\cline{2-16}
&
\multirow{2}{*}{ADD(-S)} & AUC of  & AUC of &
% \multirow{2}{*}{ADD(-S)} & AUC of  & AUC of &
\multirow{2}{*}{ADD(-S)} & AUC of  & AUC of &
\multirow{2}{*}{ADD(-S)} & AUC of  & AUC of &
\multirow{2}{*}{ADD(-S)} & AUC of  & AUC of &
\multirow{2}{*}{ADD(-S)} & AUC of  & AUC of \\
&   & ADD-S & ADD(-S) &  & ADD-S & ADD(-S) &  & ADD-S & ADD(-S) &  & ADD-S & ADD(-S) &  & ADD-S & ADD(-S)\\
\hline
\texttt{002\_master\_chef\_can}      & \na & 84.0 & 50.9 & \na & \na & \na & \na & \na & \na & 41.5 & 96.3 & 65.2 & 61.0 & 94.7 & 63.1 \\
\texttt{003\_cracker\_box}           & \na & 76.9 & 51.7 & \na & \na & \na & \na & \na & \na & 83.2 & 97.0 & 88.8 & 86.7 & 91.2 & 79.8 \\
\texttt{004\_sugar\_box}             & \na & 84.3 & 68.6 & \na & \na & \na & \na & \na & \na & 91.5 & 98.9 & 95.0 & 94.5 & 94.7 & 89.8 \\
\texttt{005\_tomato\_soup\_can}      & \na & 80.9 & 66.0 & \na & \na & \na & \na & \na & \na & 65.9 & 96.5 & 91.9 & 73.8 & 95.6 & 89.7 \\
\texttt{006\_mustard\_bottle}        & \na & 90.2 & 79.9 & \na & \na & \na & \na & \na & \na & 90.2 & 100.0 & 92.8 & 93.3 & 93.2 & 84.2 \\
\texttt{007\_tuna\_fish\_can}        & \na & 87.9 & 70.4 & \na & \na & \na & \na & \na & \na & 44.2 & 99.4 & 94.2 & 68.7 & 97.1 & 94.2 \\
\texttt{008\_pudding\_box}           & \na & 79.0 & 62.9 & \na & \na & \na & \na & \na & \na & 2.8 & 64.6 & 44.7 & 98.7 & 96.5 & 93.2 \\
\texttt{009\_gelatin\_box}           & \na & 87.1 & 75.2 & \na & \na & \na & \na & \na & \na & 61.7 & 97.1 & 92.5 & 96.0 & 97.0 & 93.6 \\
\texttt{010\_potted\_meat\_can}      & \na & 78.5 & 59.6 & \na & \na & \na & \na & \na & \na & 64.9 & 86.0 & 80.2 & 67.5 & 87.9 & 78.0 \\
\texttt{011\_banana}                 & \na & 85.9 & 72.3 & \na & \na & \na & \na & \na & \na & 64.1 & 96.3 & 85.8 & 54.7 & 90.1 & 75.4 \\
\texttt{019\_pitcher\_base}          & \na & 76.8 & 52.5 & \na & \na & \na & \na & \na & \na & 99.0 & 99.9 & 98.5 & 90.2 & 95.9 & 89.2 \\
\texttt{021\_bleach\_cleanser}       & \na & 71.9 & 50.5 & \na & \na & \na & \na & \na & \na & 73.8 & 94.2 & 84.3 & 75.7 & 92.3 & 81.9 \\
\texttt{024\_bowl$^{*}$}                  & \na & 69.7 & 69.7 & \na & \na & \na & \na & \na & \na & 37.7 & 85.7 & 85.7 & 10.7 & 73.6 & 73.6 \\
\texttt{025\_mug}                    & \na & 78.0 & 57.7 & \na & \na & \na & \na & \na & \na & 61.5 & 99.6 & 94.0 & 63.5 & 90.5 & 76.4 \\
\texttt{035\_power\_drill}           & \na & 72.8 & 55.1 & \na & \na & \na & \na & \na & \na & 78.5 & 97.5 & 90.1 & 83.7 & 94.7 & 85.0 \\
\texttt{036\_wood\_block$^{*}$}           & \na & 65.8 & 65.8 & \na & \na & \na & \na & \na & \na & 59.5 & 82.5 & 82.5 & 63.6 & 85.7 & 85.7 \\
\texttt{037\_scissors}               & \na & 56.2 & 35.8 & \na & \na & \na & \na & \na & \na & 3.9 & 63.8 & 49.5 & 46.6 & 85.3 & 70.1 \\
\texttt{040\_large\_marker}          & \na & 71.4 & 58.0 & \na & \na & \na & \na & \na & \na & 7.4 & 88.0 & 76.1 & 16.0 & 78.8 & 68.3 \\
\texttt{051\_large\_clamp$^{*}$}          & \na & 49.9 & 49.9 & \na & \na & \na & \na & \na & \na & 69.8 & 89.3 & 89.3 & 72.0 & 84.1 & 84.1 \\
\texttt{052\_extra\_large\_clamp$^{*}$}   & \na & 47.0 & 47.0 & \na & \na & \na & \na & \na & \na & 90.0 & 93.5 & 93.5 & 88.0 & 92.5 & 92.5 \\
\texttt{061\_foam\_brick$^{*}$}           & \na & 87.8 & 87.8 & \na & \na & \na & \na & \na & \na & 71.9 & 96.9 & 96.9 & 58.6 & 78.3 & 78.3 \\
\hline
\multicolumn{1}{l|}{\multirow{1}{*}{\textbf{Avg.} $\uparrow$}}            & \na & 75.9 & 61.3 & 53.9 & \na & \na & \na & 88.1 & 81.9 & 60.1 & 91.6 & 84.3 & 69.7 & 90.0 & 82.2 \\
\hline
\end{tabular}%
}
\captionsetup{font=small}
\caption*{\raggedright \textsuperscript{$\dagger$} Method is End-to-End by using a differentiable P$n$P strategy.}
\end{table*}

\section{Additional Results on SPEED+}
\label{supp:speedplus}

\begin{figure*}[h]
    \centering
    \includegraphics[width=\textwidth]{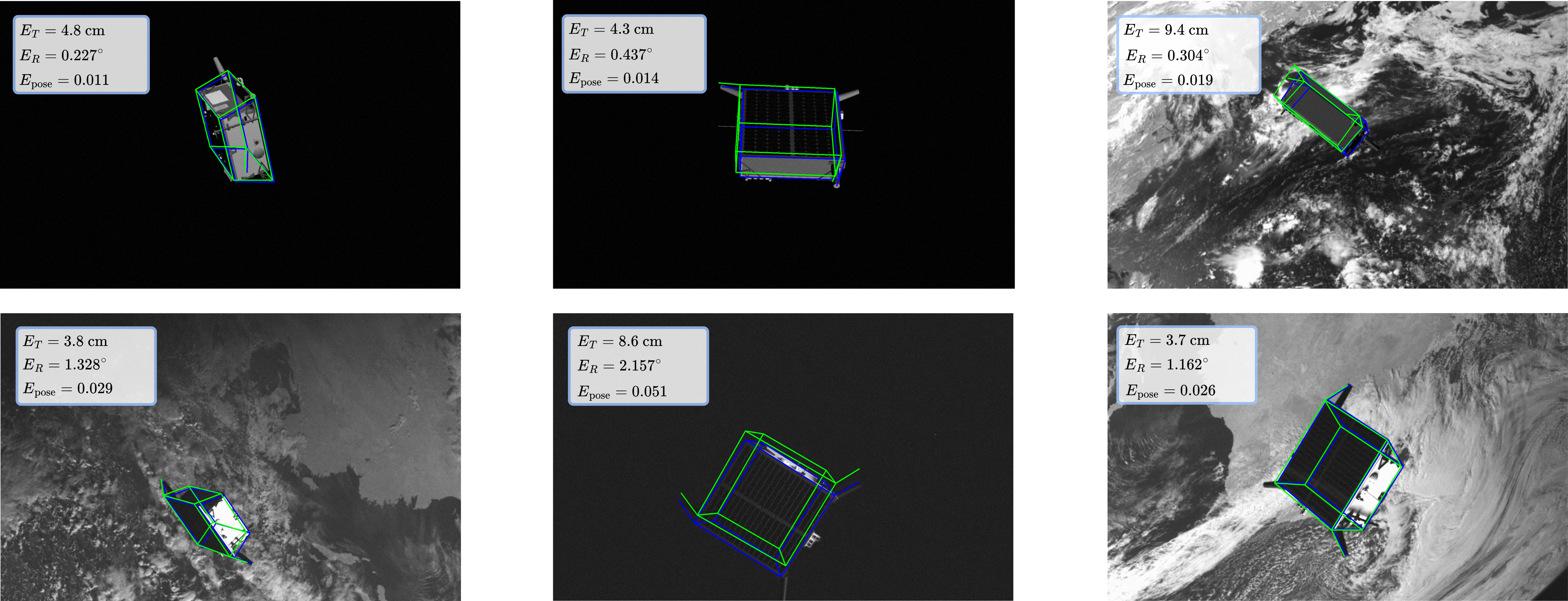}
    \caption{\textbf{Qualitative results on SPEED+.} Examples of accurate pose predictions from \methodname{} on the \texttt{synthetic} (1$^\text{st}$ row) and \texttt{lightbox} (2$^\text{nd}$ row) subsets of SPEED+~\cite{park2022speed+}. Green bounding boxes show predicted poses, while blue bounding boxes indicate ground-truth poses.}
    \label{supp:speed+qualitative}
\end{figure*}

We further evaluate \methodname{} on SPEED+~\cite{park2022speed+}, a spacecraft pose estimation dataset where P$n$P-based solutions currently predominate~\cite{park2024robust,BECHINI2025198,PARK2023640,wang2023bridging}. \methodname{} is trained on the \texttt{synthetic} train split and evaluated on its test set. To assess cross-domain generalization, we report inference results on the additional test domains of SPEED+, namely, \texttt{lightbox}, simulating diffuse orbital illumination, and \texttt{sunlamp}, replicating direct sunlight via a specialized lamp setup. As summarized in Table~\ref{supp:table_speed+}, \methodname{} surpasses SPNv2~\cite{park2024robust} across all metrics while being faster at inference. It also outperforms the P$n$P-based YOLOv8s-pose~\cite{BECHINI2025198} on the \texttt{lightbox} split and achieves superior translation accuracy on \texttt{sunlamp}. These results further support the applicability of \methodname{} to spacecraft pose estimation. Qualitative results are shown in Figure~\ref{supp:speed+qualitative}.

\begin{table}[H]
  \centering
  % smaller font
  \scriptsize
  % tighten column padding
  \setlength{\tabcolsep}{3pt}
  % reduce row height
  \renewcommand{\arraystretch}{0.5}
  \caption{\textbf{Results on SPEED+.} Performance in terms of Translation $E_T$, Rotation $E_R$ and Pose $E_{\text{pose}}$ errors as defined in~\cite{park2024robust}. The inference time of SPNv2 is measured on the same device and at the same input resolution (768$\times$512) as ours.}

  \label{supp:table_speed+}
  \resizebox{\columnwidth}{!}{%
    \begin{tabular}{@{}c l c c c@{}}
      \toprule
       & 
      & YOLOv8s-pose~\cite{BECHINI2025198}
      & $^\dagger$SPNv2~\cite{park2024robust}
      & \textbf{Ours} \\
      \cmidrule(lr){3-5}
      & Latency (ms)       &    \na    &    80.25    & \textbf{60.46} \\
      \midrule
      \multirow{3}{*}{\texttt{Synthetic}}
        & $E_T$ [m]      & \na         & \na      & \textbf{0.184} \\
        & $E_R$ [$^\circ$] & \na      & \na      & \textbf{3.462} \\
        & $E_{\text{pose}}$ & \na      & \na      & \textbf{0.089} \\
      \midrule
      \multirow{3}{*}{\texttt{Lightbox}}
        & $E_T$ [m]      & 0.758      & 0.368      & \textbf{0.300} \\
        & $E_R$ [$^\circ$] & 18.00    & 20.258     & \textbf{15.096} \\
        & $E_{\text{pose}}$ & 0.432   & 0.411      & \textbf{0.3101} \\
        \midrule
      \multirow{3}{*}{\texttt{Sunlamp}}
        & $E_T$ [m]        & 0.518  & 0.457    & \textbf{0.430} \\
        & $E_R$ [$^\circ$] & \textbf{19.80} & 34.916 & \underline{29.128} \\
        & $E_{\text{pose}}$& \textbf{0.432} & 0.682  & \underline{0.574} \\
      \bottomrule
    \end{tabular}%
  }
  \raggedright
  \scriptsize $^\dagger$Results from the direct pose regression head of SPNv2.
\end{table}

\end{document}